\documentclass[10pt,twocolumn,letterpaper]{article}

\usepackage[pagenumbers]{cvpr}
\usepackage[numbers,sort&compress]{natbib}

\usepackage[dvipsnames]{xcolor}

\definecolor{cvprblue}{rgb}{0.21,0.49,0.74}
\usepackage[pagebackref,breaklinks,colorlinks,citecolor=cvprblue]{hyperref}

\usepackage{amsmath}
\usepackage{multirow}

\title{Language Embedded 3D Gaussians for Open-Vocabulary Scene Understanding}

\author{Jin-Chuan Shi\textsuperscript{1} \quad Miao Wang\textsuperscript{1,2\text{\textbf{*}}} \quad Hao-Bin Duan\textsuperscript{1} \quad Shao-Hua Guan\textsuperscript{1}\\
\textsuperscript{1}State Key Laboratory of Virtual Reality Technology and Systems, Beihang University \\
\textsuperscript{2}Zhongguanchun Laboratory \\
{\fontsize{10}{10}\selectfont \url{https://buaavrcg.github.io/LEGaussians}}
}

\newcommand\blfootnote[1]{%
\begingroup 
\renewcommand\thefootnote{}\footnote{#1}%
\addtocounter{footnote}{-1}%
\endgroup 
}

\begin{document}

\twocolumn[{%
 \renewcommand\twocolumn[1][]{#1}%
 \maketitle
 \begin{center}
   \centering
   \includegraphics[width=\textwidth]{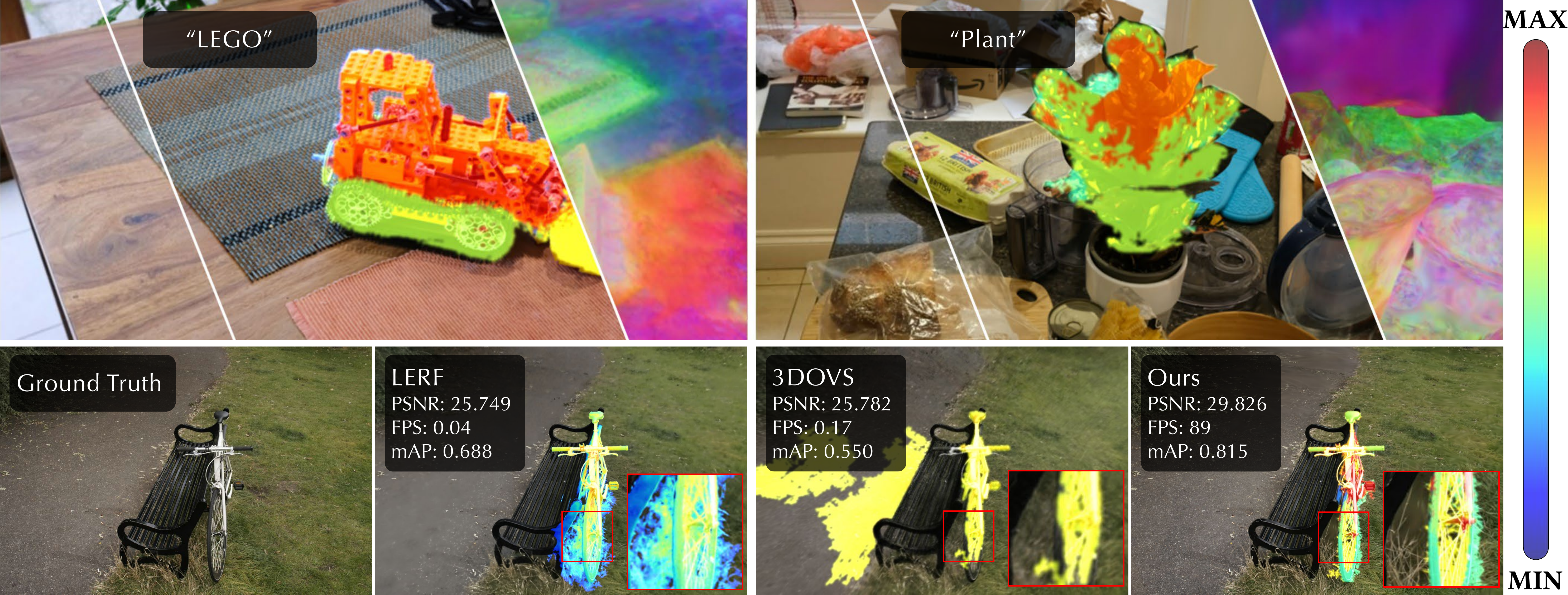}
   \captionof{figure}{
   We present Language Embedded 3D Gaussians, a novel scene representation for open-vocabulary querying. The top row visualizes the original image, novel view synthesis result with query relevancy and PCA of learned semantic features. The bottom row compares our method with other language-embedded representations. The right-side bar maps relevancy values to heatmap colors. Our method achieves better fidelity and query accuracy while rendering at higher frame rates.
   }
   \label{fig:teaser}
 \end{center}
 }]

\blfootnote{\text{\textbf{*}} Corresponding author.}

\begin{abstract}
Open-vocabulary querying in 3D space is challenging but essential for scene understanding tasks such as object localization and segmentation. Language-embedded scene representations have made progress by incorporating language features into 3D spaces. However, their efficacy heavily depends on neural networks that are resource-intensive in training and rendering. Although recent 3D Gaussians offer efficient and high-quality novel view synthesis, directly embedding language features in them leads to prohibitive memory usage and decreased performance. In this work, we introduce Language Embedded 3D Gaussians, a novel scene representation for open-vocabulary query tasks. Instead of embedding high-dimensional raw semantic features on 3D Gaussians, we propose a dedicated quantization scheme that drastically alleviates the memory requirement, and a novel embedding procedure that achieves smoother yet high accuracy query, countering the multi-view feature inconsistencies and the high-frequency inductive bias in point-based representations. Our comprehensive experiments show that our representation achieves the best visual quality and language querying accuracy across current language-embedded representations, while maintaining real-time rendering frame rates on a single desktop GPU.
\end{abstract}
\section{Introduction}
\label{sec:intro}

Neural Radiance Field (NeRFs)~\cite{zhang2020nerf++, barron2021mip, barron2022mip, mildenhall2021nerf} and 3D Gaussian Splatting~\cite{kerbl20233d} has advanced the development of efficient and high-quality 3D scene novel view synthesis from multi-view images. 
Nevertheless, these representations solely encapsulate geometric and appearance details without any semantic information.
To bridge this gap, language-embedded neural representations~\cite{kobayashi2022decomposing, kerr2023lerf} try to integrate semantic information from multi-view images into 3D scenes for open-vocabulary querying tasks, which allows intuitive interaction with large language models (LLMs)~\cite{chat-with-nerf-2023, yang2023llmgrounder} and human users, and powers broad applications including scene editing, VR/AR, autonomous driving, and robotic navigation.
Compared to traditional semantic labeling methods, language features from visual-language models like CLIP~\cite{radford2021learning} and DINO~\cite{caron2021emerging} offer more comprehensive semantic understanding capability, as they encompass objects with long-tail distribution, enhancing their suitability for real-world applications. 
However, accurately incorporating language embedding into current 3D scene representations, while maintaining their efficiency and visual quality, presents a significant challenge.

Recent techniques~\cite{kobayashi2022decomposing, kerr2023lerf, liu2023weakly} extract dense language features from multi-view 2D images and incorporate additional output branches in scene representation to predict semantic features. However, the quality of semantic features heavily relies on scene representation, and trivially expanding the output channels poses significant challenges in recovering high-precision and robust semantics of the scenes. Furthermore, while 3D Gaussian~\cite{kerbl20233d} are efficient and fast for high-quality 3D scene representation, embedding raw semantic information into a massive number of points can cause prohibitive memory requirements and significantly lower the efficiency of both optimization and rendering.

In this paper, we introduce Language Embedded 3D Gaussians, a semantic scene representation framework that provides both high precision and efficiency for open-vocabulary query tasks. To avoid the substantial memory requirement that would result from directly increasing the feature channels for raw semantic features in the 3D Gaussians, we employ a novel feature quantization approach that significantly decreases the computational and memory cost. By leveraging the redundancy nature of local semantic features, we construct more streamlined language features that are compactly stored on the 3D Gaussians. Additionally, we address the issue of semantic ambiguity caused by visual inconsistency from multi-view images by implementing a mechanism that lowers the spatial frequency of semantic features, guided by learned uncertainty values. This technique enables language features in the 3D Gaussians to be much smoother yet still precise.

Our extensive experiments demonstrate that our method achieves state-of-the-art quality in both novel view synthesis and open-vocabulary querying tasks, while allowing real-time rendering on consumer-level devices. 

In summary, our contributions include:
\begin{itemize}[itemsep=0pt,parsep=0pt,topsep=2bp]
    \item We introduce a novel quantization scheme that efficiently compresses and integrates semantic features into dense 3D Gaussians, ensuring efficient optimization and rendering on consumer devices while maintaining accurate semantic embedding.
    \item We propose a mechanism that leverages spatial position and semantic uncertainty of 3D Gaussians to address the semantic ambiguity arising from visual inconsistency across views.
    \item Our method outperforms other language-embedded 3D representations, delivering state-of-the-art results in visual quality, language query precision, and rendering speed at the same time.
\end{itemize}

\section{Related Work}
\label{sec:related_work}

\paragraph{Neural Rendering.}
NeRF~\cite{mildenhall2021nerf} has demonstrated superior novel view synthesis quality over traditional methods~\cite{snavely2006photo, goesele2007multi, chaurasia2013depth, eisemann2008floating, hedman2018deep}, and various efforts have been made to enhance its performance. Although many methods focus on improving NeRF's rendering quality~\cite{zhang2020nerf++, barron2021mip, barron2022mip, martin2021nerf, Tancik_2022_CVPR, rematas2022urf}, they still suffer from slow training and rendering speed.  
On the other hand, explicit and hybrid scene representations~\cite{chen2022tensorf, fridovich2022plenoxels, muller2022instant, xu2022point, barron2023zip, sun2022direct, liu2020neural}. 
These methods typically utilize hash grids~\cite{muller2022instant} and point clouds~\cite{xu2022point} to reduce computational cost of large neural networks.
Recent 3D Gaussian~\cite{kerbl20233d} sets a new standard on both rendering quality and training speed, by using fast rasterization of 3D Gaussians to replace differential volume rendering. Nevertheless, directly embedding language embeddings on the dense 3D Gaussians is non-trivial, as the massive amount of 3D points requires prohibitive memory usage and causes a drastic performance drop in both training and rendering.

\paragraph{Language Embedded Scene Representation.}
Incorporating specific semantics into NeRF's implicit MLP-based representation is difficult due to challenges in accurately identifying 3D regions. Various approaches~\cite{zhi2021place, kobayashi2022decomposing, siddiqui2023panoptic, fu2022panoptic, wang2022dm, tschernezki2022neural, kerr2023lerf, engelmann2023open} have tried to integrate semantic data into NeRF, often by adding new network branches. For instance, DFF~\cite{kobayashi2022decomposing} embeds a branch for language predictions and uses a pre-trained encoder for supervision, while LERF~\cite{kerr2023lerf} assimilates CLIP~\cite{radford2021learning} features from multi-scale image crops into hash grid-represented 3D scenes. 3D-OVS~\cite{liu20233d} takes a different route, creating dual decomposed 3d tensors for geometry and semantics, but it requires predefined segmented classes and doesn't support arbitrary queries. Other methods~\cite{kundu2022panoptic, zhang2023nerflets} decompose scenes into smaller MLPs representing local semantics. However, the quality of semantic embeddings and the efficiency of training and rendering of these methods, are all constrained by the limitations of their scene representation models.

\paragraph{Open Vocabulary 3D Scene Understanding.}
Recent progress in open vocabulary scene understanding has been marked by the integration of 2D Vision-Language Models with 3D point cloud processing~\cite{zhang2021pointclip, Zhu2022PointCLIPV2, huang2022clip2point, xue2022ulip, xue2023ulip2}. These approaches focus on aligning features and projecting 3D data into 2D to enhance zero-shot learning capabilities. Additionally, advancements in 3D object detection and segmentation~\cite{lu2023open, ding2022language, Peng2023OpenScene, zhang2023clipfo3d, takmaz2023openmask3d} have shown the effectiveness of merging point cloud data with visual features extracted from image for scene analysis~\cite{wu2023open}. However, these methods mainly address the comprehension and analysis of existing scene representations, like point clouds, rather than optimizing scene representation from multi-view images, which is the focus of our work.

\begin{figure*}[t] \centering
    \includegraphics[width=0.95\textwidth]{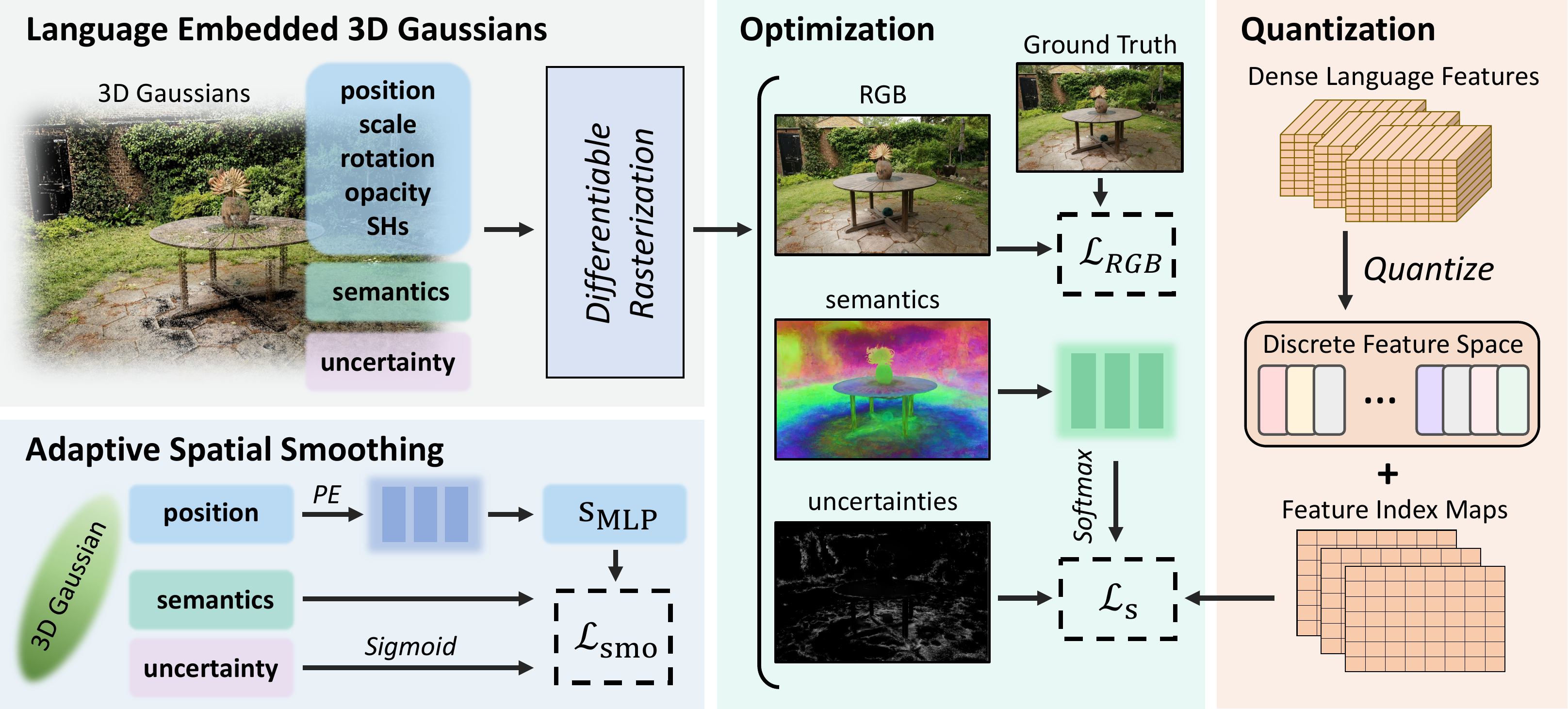}
    \caption{
    The training process for Language-embedded 3D Gaussians starts with initializing scenes following 3D Gaussian Splatting~\cite{kerbl20233d} and randomly initializing semantic features and setting uncertainty to zero. Dense language features from multi-view CLIP~\cite{radford2021learning} and DINO~\cite{caron2021emerging} are quantized to create a discrete feature space and semantic indices. These attributes of the 3D Gaussians are then rendered into 2D maps using a differentiable rasterizer. The optimization is achieved through semantic and adaptive spatial smoothing loss.
    } 
    \label{fig:figure2}
    \vspace{-3mm}
\end{figure*}

\section{Method}
\label{sec:method}
In this section, we introduce our training process of Language Embedded 3D Gaussians, including (1) a recap of 3D Gaussian Splatting~\cite{kerbl20233d} (Sec.~\ref{sec:recap}), (2) extracting dense language features from multi-view images (Sec.~\ref{sec:extraction}), (3) a quantization scheme for high-dimensional language features that creates a compact feature space for the embedding process (Sec.~\ref{sec:quantization}), (4) an embedding mechanism that alleviates semantic ambiguity caused by visual difference across views, by lowering the spatial frequency of semantic features with the learned uncertainty values (Sec.~\ref{sec:lem}).

\subsection{Recap: 3D Gaussian Splatting}
\label{sec:recap}
3D Gaussian Splatting~\cite{kerbl20233d} renders complex scenes by merging a multitude of colored 3D Gaussians, which are subsequently projected onto camera views through a rasterization process.
By employing differentiable rendering and gradient descent, the attributes of these 3D Gaussians, including position $p$, covariance $\Sigma$, color $c$, and opacity $\alpha$, are optimized to represent the 3D scene based on a collection of input images.
The result image $I$ is rendered from a specific camera pose $p_{\text{cam}}$, using the differentiable rasterization $R$, represented as:
\begin{equation}
  I = R(p, \Sigma, c, \alpha; p_{\text{cam}}) .
  \label{eq:render}
\end{equation}

\subsection{Dense Language Feature Extraction}
\label{sec:extraction}

We first extract pixel-level dense language features from visual-language models.
While CLIP~\cite{radford2021learning} encodes images into global semantic features, its direct application is not feasible for our purposes as we require pixel-level targets to learn 3D scene representations from multi-view images.
Prior studies~\cite{kerr2023lerf, liu20233d} overcome this limitation by computing multi-scale dense CLIP features for each image.
To obtain dense language embeddings from multi-view images, we employ a slightly different hierarchical random cropping technique to extract CLIP features, similar to 3DOVS~\cite{liu20233d}. 
We aggregate features from all layers and normalize them to produce the final language embeddings.

Nevertheless, features extracted from CLIP only provide a rough boundary of different semantic regions, resulting in ambiguities and inaccuracies in the language embedding of 3D scenes.
Conversely, DINO~\cite{caron2021emerging} exhibits autonomous object decomposition without requiring labeled data~\cite{amir2021deep}, as several studies~\cite{kobayashi2022decomposing, kerr2023lerf, liu20233d, wang2023autorecon} have shown that DINO effectively enhances language feature grouping without reliance on labels or prior knowledge. Therefore, we also extract DINO features as complementary to enhance the details of extracted language features.

We then concatenate the dense CLIP and DINO features extracted from multi-view images to form hybrid language feature maps.
We denote the features at the position $(x, y)$ on the CLIP and DINO feature maps of image $I$ as $ \mathbf{F}^{\text{CLIP}}_{I, x, y} \in \mathbb{R}^{d_{\text{CLIP}}} $ and $ \mathbf{F}^{\text{DINO}}_{I, x, y} \in \mathbb{R}^{d_{\text{DINO}}} $, respectively, where $ d_{\text{CLIP}} $ and $ d_{\text{DINO}} $ represent the dimension of each feature. The hybrid language feature $ \mathbf{F}_{I, x, y} \in \mathbb{R}^{d} $ is given by: 
\begin{equation}
  \mathbf{F}_{I, x, y} = \mathbf{F}^{\text{CLIP}}_{I, x, y} \oplus \mathbf{F}^{\text{DINO}}_{I, x, y} ,
  \label{eq:concat}
\end{equation}
 where $d$ is the total channels in the hybrid features.

\subsection{Quantization of Language Features}
\label{sec:quantization}

Direct integration of our extracted language features as in prior studies~\cite{kobayashi2022decomposing, kerr2023lerf, liu20233d} would be infeasible for 3D Gaussians, due to the substantial memory demands of recording high-dimensional language embeddings on them, which significantly hinders rendering efficiency and limits the maximum number of 3D Gaussians that can be trained at the same time.
To alleviate such storage and computation costs of raw language embeddings, our insight is to leverage the inherent redundancy in language features, as those within a single object share very similar semantics. Additionally, the semantics of a single scene merely cover only a small fraction of the original CLIP feature space, leading to unnecessary data stored in the 3D representation for all the queries we would need.
To this end, we propose a dedicated quantization scheme to effectively compress the semantic features extracted from multiple viewpoints, resulting in a more efficient and compact representation of scene-aware semantic features.

Specifically, our goal is to transform the hybrid language features $\mathbf{F} \in \mathbb{R}^{d}$ into a quantized version $\hat{\mathbf{F}} \in \mathbb{R}^{d}$, which approximates $\mathbf{F}$. To achieve this, we develop a discrete language feature space $\mathcal{S} = \{ \mathbf{f}_i \in \mathbb{R}^d | i=1,2,\cdots, N \}$, and use an integer index $m \in \{ 1,2,\cdots,N \}$ for retrieving the nearest language feature in $\mathcal{S}$. The quantized language feature is then given by:
\begin{equation}
\hat{\mathbf{F}} = \sum_i^N \mathbf{f}_i \cdot \text{onehot}(m)_i .
\label{eq:quant}
\end{equation}
Here, $\mathbf{f}_i$ represents the $i$-th basis of the language features, and $\text{onehot}(m) \in \mathbb{R}^N$ is the one-hot vector for index $m$. This quantization scheme effectively removes excessive information in the original language embeddings by compressing the original continuous language feature space into discrete bases, with the compression rate adjustable through the size of space $N$.

\paragraph{Quantization.}
To quantitize high-dimentinal language feature $\mathbf{F}$, we execute a max-similarity search within space $ \mathcal{S} $, utilizing cosine similarity $ \cos\langle \cdot \rangle $ as the distance metric:
\begin{equation}
    \mathcal{D}(\mathbf{F}, \mathbf{f}_i) = cos \langle \mathbf{F}^{\text{CLIP}} \cdot \mathbf{f}_i^{\text{CLIP}} \rangle 
     + \lambda_{\text{DINO}} cos \langle \mathbf{F}^{\text{DINO}} \cdot \mathbf{f}_i^{\text{DINO}} \rangle ,
  \label{eq:q_dist}
\end{equation}
where $\lambda_{\text{DINO}}$ serves as a hyper-parameter that modulates the significance of DINO within the semantic feature.

Following VQ-VAE~\cite{van2017neural}, the selected language feature index from the set $\mathcal{S}$ is determined as $m = {\text{argmax}}_{i}(\mathcal{D}(\mathbf{F}, \mathbf{f}_i))$, and the quantization of $ \mathbf{F} $ is computed using Eq.~(\ref{eq:quant}).
The result for each image after this quantization procedure is a semantic indices map, denoted as $\mathcal{M} \in \mathbb{R}^{H \times W \times 1}$.

\vspace{-2.0mm}
\paragraph{Optimization.}
During the quantization of all language features extracted from multi-view images, the optimization of the discrete feature space $\mathcal{S}$ is simultaneously accomplished by minimizing the cosine similarity loss between the language features $\mathbf{F}_i$ and the quantization $\hat{\mathbf{F_i}}$:
\vspace{-1.0mm}
\begin{equation}
\begin{aligned}
    \mathcal{L}_{cos}(\mathbf{F}_i) = 
    & (1 - cos \langle \mathbf{F}_i^{\text{CLIP}} \cdot \hat{\mathbf{F}}_i^{\text{CLIP}} \rangle) \\
    & + \lambda_{\text{DINO}} (1 - cos \langle \mathbf{F}_i^{\text{DINO}} \cdot \hat{\mathbf{F}_i^{\text{DINO}}} \rangle) .
  \label{eq:q_loss_cos}
\end{aligned}
\vspace{-1.0mm}
\end{equation}

Furthermore, to prevent quantization collapse and ensure maximal utilization of each feature in the feature space, we have devised a load balancing loss inspired by the design of the Switch Transformer~\cite{fedus2022switch}. 
The load balancing loss is calculated by performing an element-wise multiplication of the utilization ratio $\mathbf{r} \in \mathbb{R}^N$ and the mean selection probability $\mathbf{p} \in \mathbb{R}^N$ of each feature, followed by their summation:
\vspace{-1.0mm}
\begin{equation}
    \mathcal{L}_{lb} = \sum^N(\mathbf{r} \circ \mathbf{p})
  \label{eq:q_loss_lb} ,
\vspace{-1.0mm}
\end{equation}
where $\circ$ represents the element-wise product.
We provide a detailed description of the load balancing loss in the supplementary material.

In summary, when quantizing the multi-view semantic features, we optimize the discrete feature space $\mathcal{S}$ as well as the semantic indices maps $\mathcal{M}$ using the following loss:
\begin{equation}
    \mathcal{L}_q = \lambda_{cos} \mathcal{L}_{cos} + \lambda_{lb} \mathcal{L}_{lb}
  \label{eq:q_loss_cb} .
\end{equation}

\subsection{Language Embedded 3D Gaussians}
\label{sec:lem}

Utilizing the discrete feature space $\mathcal{S}$ and index maps $\mathcal{M}$ in the previous section, we embed the compressed language features into 3D Gaussians for open-vocabulary scene understanding. Concretely, we expand the number of channels of each 3D Gaussian to include a compact feature vector representing the discrete index of language feature bases. To address semantic ambiguity arising from visual disparities across various viewpoints, we introduce a novel mechanism to reduce the spatial frequency of language embeddings through an adaptive learning loss based on the learned uncertainty values on each point.

\subsubsection{Compact Semantic Features on 3D Gaussians}
The inherent attributes of 3D Gaussians, such as color, are rendered onto the screen through the processes of rasterization and alpha blending (see Sec. \ref{sec:recap}). 
However, embedding discrete semantic indices $m$ onto 3D Guassians would lead to erroneous results when optimizing through differential rendering, as these indices are not in a continuous space. 
Instead of directly embedding indices, we learn another continuous and compact semantic feature vectors, denoted as $\mathbf{s}_G \in \mathbb{R}^{d_{s}}$, where $d_{s}$ is a hyper-parameter for controlling the storage capability for semantics. 
We then render these compact semantic feature vectors into a 2D feature map with rasterization and alpha blending, and decode the 2D feature map into the discrete semantic indices $m$ using a tiny MLP decoder:
\begin{equation}
    \hat{\mathcal{M}} = \text{softmax}(D_{\text{MLP}}(R_s(\mathcal{G};p_{\text{cam}}))) 
  \label{eq:lem_dec} ,
\end{equation}
where $ R_s(\mathcal{G};p_{\text{cam}}) \in \mathbb{R}^{H \times W \times d_{s}} $ represents the semantic feature rendering from a set of 3D Gaussians $ \mathcal{G} $ viewed from camera pose $p_{\text{cam}}$. $D_{\text{MLP}}$ represents the tiny MLP decoder. During training process, a softmax operation is applied to the decoder's output, yielding the semantic feature index distribution $ \hat{\mathcal{M}} \in \mathbb{R}^{H \times W \times N} $, where $ H $ and $ W $ denote the height and width of the image, respectively.
To optimize the semantic features of the 3D Gaussians and the MLP decoder, we apply the cross-entropy loss:
\begin{equation}
    \mathcal{L}_{\text{CE}} = \text{CE}(\hat{\mathcal{M}}, \mathcal{M})
  \label{eq:lem_loss_ce0} ,
\end{equation}
where $ \mathcal{M} \in \mathbb{R}^{H \times W \times 1} $ denotes the discrete semantic indices map of the image extracted during the language feature quantization process (Sec.~\ref{sec:quantization}).

\begin{table*}[t]\centering
    \resizebox{\textwidth}{!}{
    \begin{tabular}{*{12}{c}}
        \toprule
       Method & PSNR$\uparrow$ & SSIM$\uparrow$ & LPIPS$\downarrow$ & mPA$\uparrow$ & mP$\uparrow$ & mIoU$\uparrow$ & mAP$\uparrow$ & FPS$\uparrow$ & Memory$\downarrow$ & Storage$\downarrow$ & Training Time$\downarrow$\\
        \midrule
        DFF \cite{kobayashi2022decomposing} & 25.378 & 0.712 & 0.312 & 0.817 & 0.124 & 0.091 & 0.199 & 0.202 & 42GB+14GB & 41GB & 184min \\
        LERF \cite{kerr2023lerf} & 25.749 & 0.811 & 0.317 & 0.890 & 0.475 & 0.403 & 0.688 & 0.04 & 25GB+6GB & 320MB & $\mathbf{54min}$ \\
        3DOVS \cite{liu20233d} & 25.782 & 0.733 & 0.295 & 0.905 & 0.529 & 0.458 & 0.550 & 0.17 & 57GB+15GB & 205GB & 158min \\
        Ours & $\mathbf{29.826}$ & $\mathbf{0.901}$ & $\mathbf{0.112}$ & $\mathbf{0.947}$ & $\mathbf{0.753}$ & $\mathbf{0.578}$ & $\mathbf{0.815}$ & $\mathbf{89}$ & $\mathbf{11GB}$+$\mathbf{12GB}$ & $\mathbf{15MB}$ & 68min \\
        \bottomrule
    \end{tabular}
    }
    \caption{ Quantitative comparison of our method with DFF \cite{kobayashi2022decomposing}, LeRF \cite{kerr2023lerf}, 3DOVS \cite{liu20233d}.}
    \label{tab:comp}
    \vspace{-3mm}
\end{table*}

\subsubsection{Semantic Feature Smoothing}

Due to viewing angles, illuminations, existence of specular and semi-transparent materials, language features of the same spatial location in multi-view images may exhibit high variance, posing a challenge for precise semantic learning in 3D scenes. 
The high variance in language features at the same spatial location in multi-view images, due to differences in viewing angles, illumination, and the presence of specular and semi-transparent materials, presents challenges in accurately learning semantics in 3D scenes. Moreover, partially occluded objects may have their semantic features inaccurately extracted due to the lack of detection of their entirety.
When these biased semantic features undergo quantization, the error may be amplified, resulting in a single 3D position being linked to several distinct language feature bases and, therefore, different discrete feature indices.
To this end, we introduce a smoothing strategy that limits the spatial frequency of semantic features on 3D Gaussians. This is achieved by incorporating an adaptive loss based on a learnable uncertainty value for each 3D Gaussian.

\paragraph{Learning of uncertainty.}
To represent the variance associated with the semantic feature for each Gaussian, we record an optimizable semantic uncertainty on each point, denoted as $u \in [0,1]$. A higher $u$ suggests that the semantic feature may exhibit instability and undergo frequent changes during the optimization process.
The uncertainty values are jointly optimized when training the compact semantic features $\mathbf{s}_G$:
\begin{equation}
    \mathcal{L}_{\text{CE}} = 
    \frac
{\sum\text{CE}(\hat{\mathcal{M}}, \mathcal{M}) \circ (1 - R_u(\mathcal{G};p_{\text{cam}}))}
    {H \times W}
  \label{eq:lem_loss_ce1} ,
\end{equation}
where 
$R_u(\mathcal{G};p_{\text{cam}}) \in \mathbb{R}^{H \times W \times 1} $ is the rendered 2D uncertainty map from camera pose $p_{\text{cam}}$.
At the same time, we regularize these uncertainty values to avoid converging to a trivial solution where all 3D Gaussians have the maximum uncertainties:
\begin{equation}
    \mathcal{L}_{u} = 
    \frac
    {\sum R_u(\mathcal{G};p_{\text{cam}})}
    {H \times W}
  \label{eq:lem_loss_uncer} .
\end{equation}
In summary, the total semantic loss for optimizing the compact semantic features of the 3D Gaussians and the MLP decoder is defined as: 
\begin{equation}
    \mathcal{L}_{s} = \lambda_{\text{CE}} \mathcal{L}_{\text{CE}} + \lambda_{u} \mathcal{L}_{u},
  \label{eq:lem_loss}
\end{equation}
where $\lambda_{u}$ is a hyper-parameter for controlling the weight of regularization. We initialize the uncertainty values to zero at the start of training, and inconsistent semantic feature indices will lead to increased $u$ during optimization.

\paragraph{Adaptive spatial smoothing loss.}

Building on the earlier observation that spatially adjacent positions typically display similar semantic features, we utilize this prior knowledge to deliberately decrease the spatial frequency of the embedded compact semantic features in 3D Gaussians, particularly for those with high uncertainty values. 
We harness the inductive biases of coordinate-based MLPs, known for learning low-frequency representations of target signals, to regularize the semantic features in 3D Gaussians. We calculate the smoothed semantic features $\mathbf{s}_\text{MLP} \in \mathbb{R}^{d_{s}}$ by inputting the position of each 3D Gaussian into a small MLP:
\begin{equation}
    \mathbf{s}_{\text{MLP}} = \text{MLP}(\text{PE}(\mathbf{p}))
  \label{eq:lem_fmlp} ,
\end{equation}
where $\mathbf{p} \in \mathbb{R}^{3}$ represents the position of Gaussian, and $\text{PE}$ represents the positional encoding used in NeRF~\cite{mildenhall2021nerf}. We set a low frequency in the positional encoding to encourage spatial smoothness.
We then apply the following loss for imposing the spatial smoothness regularization, where the degree of smoothness is adaptively controlled based on the learned uncertainty values:
\begin{equation}
    \mathcal{L}_{\text{smo}} = \Vert \mathbf{s}_{\text{MLP}} - \mathbf{s}_G^* \Vert_2 
    + \max(\mathbf{u}_G^*, w_s) \Vert \mathbf{s}_{\text{MLP}}^* - \mathbf{s}_G \Vert_2
  \label{eq:lem_loss_mlp} ,
\end{equation}
where the $*$ denotes the stop gradient operator, and $w_s$ is the minimal weight for the semantic smoothing term.

To summarize, the combined loss for optimization of semantic features, semantic uncertainties, and MLPs is given by: 
\begin{equation}
    \mathcal{L} = \lambda_s \mathcal{L}_s + \lambda_{\text{smo}} \mathcal{L}_{\text{smo}}
  \label{eq:loss} .
\end{equation}

\section{Implementation Details}
\label{sec:implementation}
We implement our method using PyTorch~\cite{paszke2017automatic}, and incorporate the CUDA kernel from 3D Gaussian Splatting~\cite{kerbl20233d} to speed up the rasterization rendering process. 
Our method optimizes the scene's geometry and appearance with the same RGB loss following 3D Gaussian Splatting and enables adaptive density control of 3D Gaussians during the reconstruction process. 
We modify the CUDA kernel to enable the rendering of semantic features on the 3D Gaussians, and ensure that the optimizing of these semantic parameters does not affect original reconstruction quality.
We set $d_{s} = 8$ and $w_S = 0.1$ in our modified 3D Gaussians, and set $\lambda_{D} = \lambda_{lb} = 0.5$, and all other $\lambda$ values are set to $1$.
After the phase of extracting dense semantic features, which takes about 30 minutes, our model can be trained on one RTX3090 GPU for about 1 hour.
The training involves 30,000 iterations, utilizing the Adam optimizer~\cite{loshchilov2019decoupled}, with a learning rate set to $0.001$ and betas equal to $(0.9, 0.999)$.
We leave the inference details of open-vocabulary language querying and semantic relevancy calculation in the supplementary material.
\section{Experiments}
\label{sec:experiments}

\begin{figure*}[t] \centering
    \includegraphics[width=\textwidth]{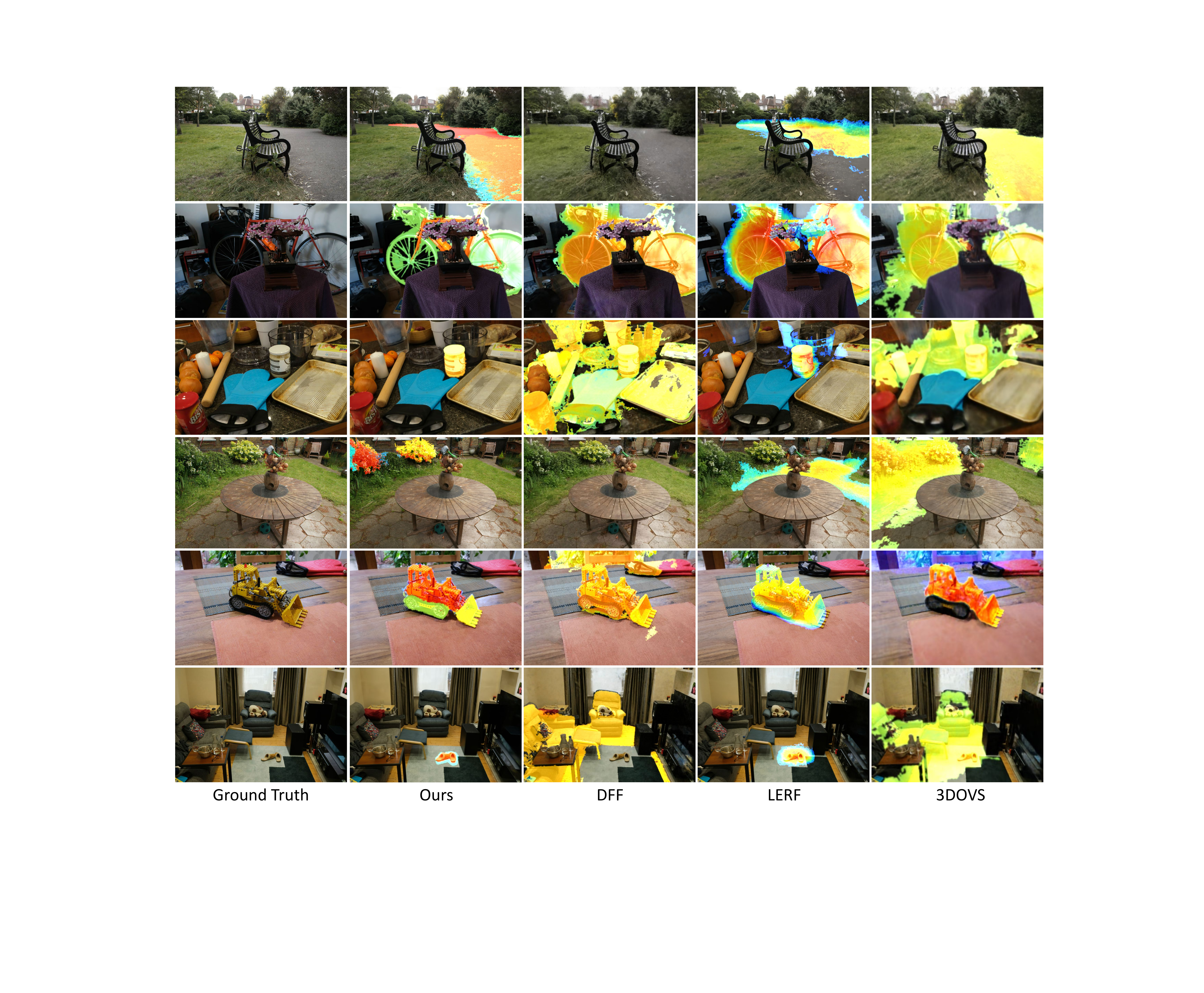}
    \caption{
    Comparison of novel view synthesis and query relevance visualization. Left to right: Ground truth novel view synthesis, novel view images with relevance visualization from our method, DFF~\cite{kobayashi2022decomposing}, LeRF~\cite{kerr2023lerf}, and 3DOVS~\cite{liu20233d}. Top to bottom: Query words ``asphalt ground'', ``bicycle'', ``jar of coconut oil'', ``flower'', ``LEGO Technic 856 Bulldozer'', and ``brown shoes''.
    }
    \label{fig:comp}
    \vspace{-5mm}
\end{figure*}

\begin{table}[t]\centering
    \resizebox{1.0\columnwidth}{!}{
    \begin{tabular}{*{6}{c}}
        \toprule
        & & mPA$\uparrow$ & mP$\uparrow$ & mIoU$\uparrow$ & mAP$\uparrow$ \\
        \midrule
        \multirow{2}{*}{Quantization}
        & w/o DINO               & 0.927 & 0.604 & 0.481 & 0.676 \\
        & w/o $\mathcal{L}_{lb}$ & 0.939 & 0.666 & 0.544 & 0.738 \\
        \midrule
        \multirow{3}{*}{Embedding}
        & w/o $u_G$  & 0.944 & 0.717 & 0.576 & 0.773 \\
        & w/o MLP    & 0.945 & 0.715 & $\mathbf{0.580}$ & 0.774 \\
        & w/o $u_G$ $\&$ MLP & 0.944  & 0.717 & $\mathbf{0.580}$ & 0.774 \\
        \midrule
        \multirow{1}{*}{Ours}
        & -  & $\mathbf{0.947}$ & $\mathbf{0.753}$ & 0.578 & $\mathbf{0.815}$ \\
        \bottomrule
    \end{tabular}
    }
    \caption{Quantitative results of ablation experiments.}
    \label{tab:ab}
    \vspace{-3mm}
\end{table}

\begin{figure}[t] \centering
\includegraphics[width=0.48\textwidth]{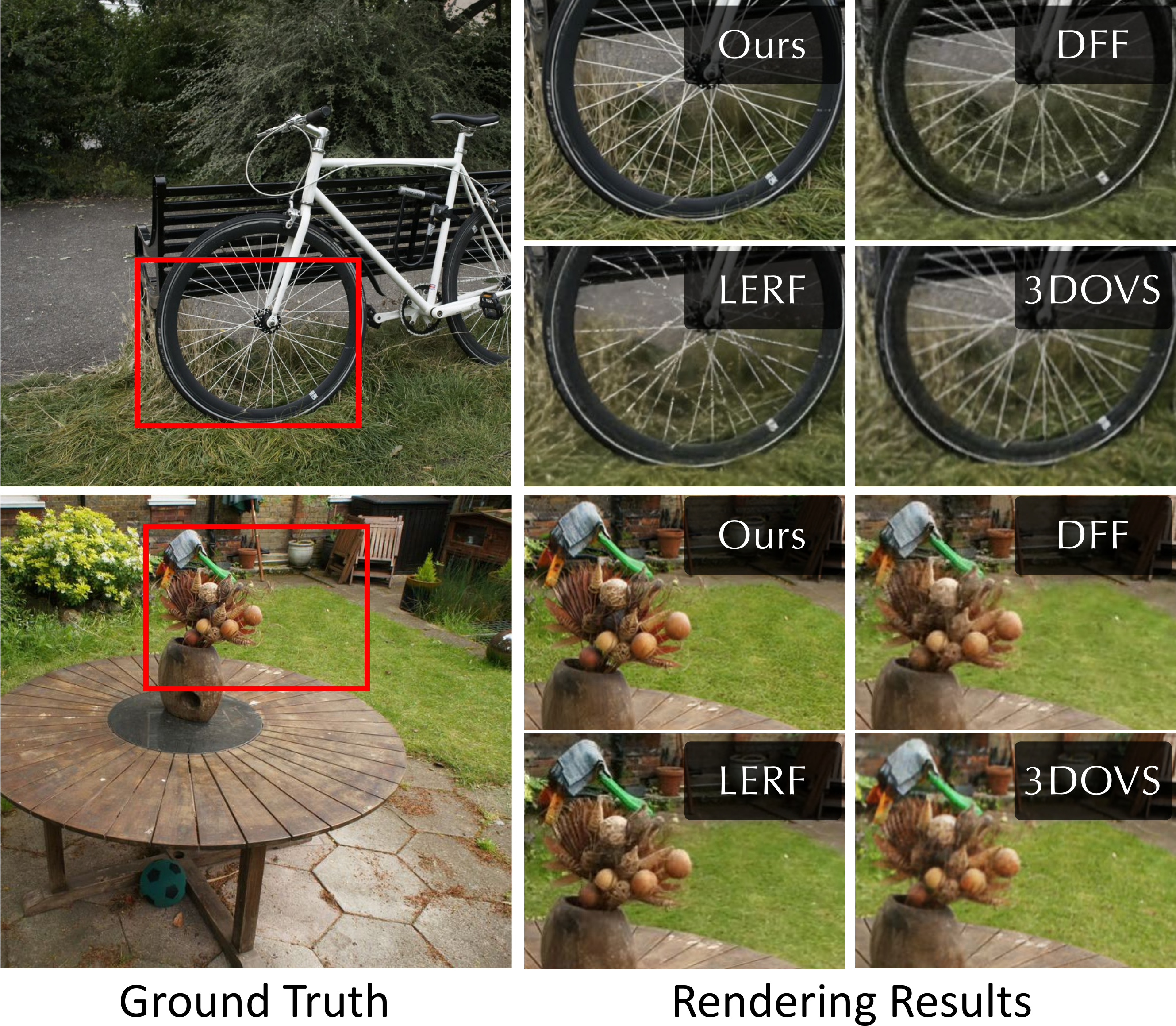}
    \caption{ Visual quality comparison of novel view synthesis results. Our method is able to recover more detailed geometry and appearance compared to other methods.
    } 
    \label{fig:render}
    \vspace{-3mm}
\end{figure}

\begin{figure}[t] \centering
\includegraphics[width=0.48\textwidth]{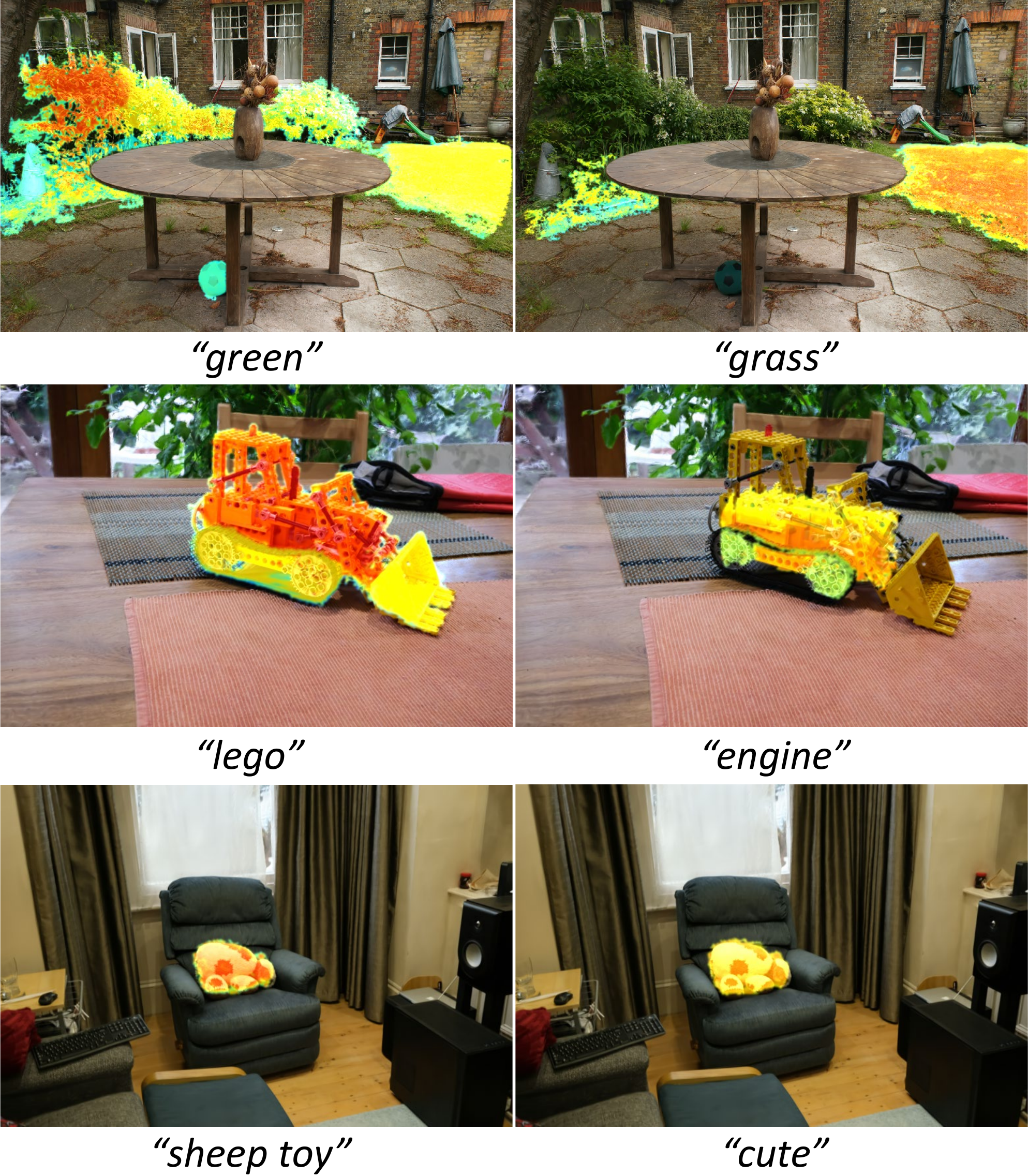}
    \caption{ Images of various open-vocabulary queries. 
    } 
    \label{fig:open}
    \vspace{-3mm}
\end{figure}

\begin{figure}[t] \centering
\includegraphics[width=0.48\textwidth]{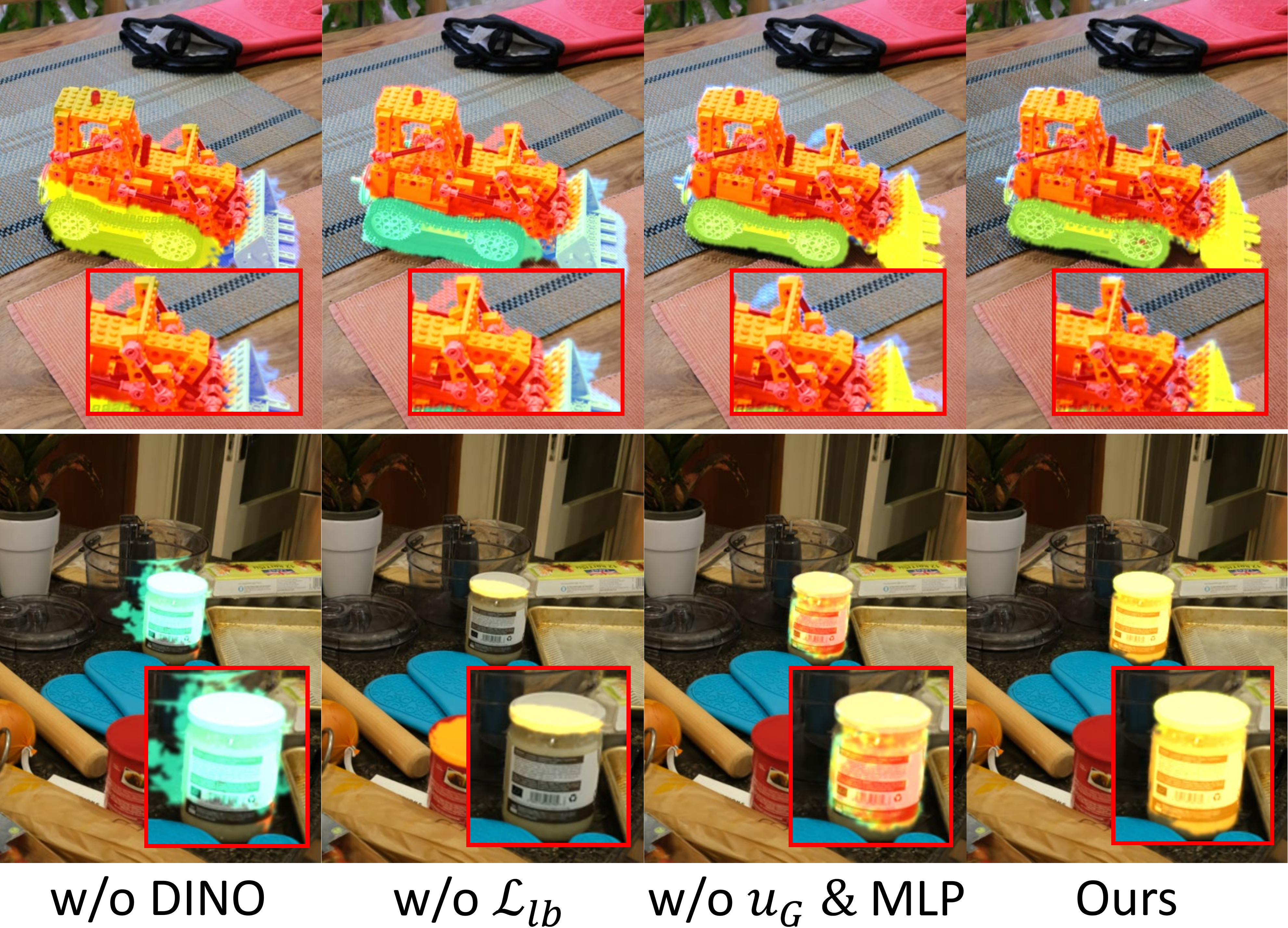}
    \caption{ Comparison of ablation experiments. 
    } 
    \label{fig:abl}
    \vspace{-3mm}
\end{figure}

\subsection{Basic Setups}
\paragraph{Dataset.}
For a simultaneous evaluation of visual and semantic embedding quality, we select six scenes (excluding Stump) from the Mip-NeRF360 dataset~\cite{barron2022mip} and manually annotate segmentation maps for each scene in the evaluation set. 
Each scene encompasses 180 to 320 images captured from various angles, and the evaluation set is chosen randomly with many novel-view images. 
Segmentation masks are annotated for primary objects in each scene.

\paragraph{Baseline Methods and Metrics.}
We conduct a comparative evaluation of our method with DFF~\cite{kobayashi2022decomposing}, LeRF~\cite{kerr2023lerf}, and 3DOVS~\cite{liu20233d}, focusing on visual quality, language-embedded accuracy, rendering speed and model efficiency. 
To measure the visual quality in novel views, we report the PSNR, SSIM, and LPIPS~\cite{zhang2018unreasonable} metrics. 
For the accuracy of language embedding, we measure the mean intersection
over union (mIoU), mean pixel accuracy (mPA), mean precision (mP), and mean average precision (mAP) based on our annotations.
The rendering speed (FPS) is measured by rendering the images with language features at a consistent resolution.
Additionally, model efficiency is evaluated based on CPU and GPU memory usage during training, as well as data storage requirements and training duration.

\subsection{Comparisons}
We compare our method both qualitatively and quantitatively with DFF~\cite{kobayashi2022decomposing}, LeRF~\cite{kerr2023lerf}, and 3DOVS~\cite{liu20233d} on our annotated Mip-NeRF360 dataset using a single RTX3090 GPU, following their default parameters but at the same resolution as our method.

\vspace{-2.0mm}
\paragraph{Qualitative Results.}
Fig.~\ref{fig:comp} displays a qualitative comparison of the novel view synthesis and semantic embedding results, demonstrating our method's efficacy in querying challenging objects in both indoor and outdoor scenes. 
Our approach notably delivers the highest visual rendering quality and query accuracy across all the tested scenes. 
Specifically, DFF~\cite{kobayashi2022decomposing} fails to identify "asphalt ground" in scene "bicycle" and "flower" in scene "garden". 
This may be caused by its use of LSeg~\cite{li2022languagedriven}, which is unstable to compute correct features in complex scenes. 
Moreover, due to predetermined query categories during training, 3DOVS~\cite{liu20233d} shows poor performance in scenes with complex objects.
While LERF~\cite{kerr2023lerf} can locate queried objects, its grid-based scene representation limits its ability to define clear boundaries. In contrast, our point-based approach supports high-frequency embedded semantic features, allowing for enhanced spatial semantic accuracy simply by incorporating more 3D Gaussians into scenes. This is facilitated by our quantization method, which substantially lowers memory costs. Additionally, our adaptive semantic smoothing technique effectively manages the high variance and ambiguity observed from different viewpoints, selectively applying spatial consistency as required.

\paragraph{Quantitative Results.}

Tab.~\ref{tab:comp} presents a comparison across various metrics, including novel view synthesis quality, open-vocabulary query accuracy, and computational efficiency. We report both host memory and video memory usage, as well as the disk space used for storing the learned language features. Our approach outperforms others in rendering quality and semantic query accuracy, while also offering lower computational demands and a significant speed increase, nearly 100 times faster in inference. It's noteworthy that methods like 3DOVS and DFF require extensive memory and storage due to their use of raw language features during training. In contrast, our use of the quantization scheme facilitates the incorporation of detailed semantics into complex 3D scenes with numerous 3D Gaussians, and concurrently achieves the most efficient storage utilization among the baseline methods.

\subsection{Open-vocabulary Query}

Fig.~\ref{fig:open} illustrates the results of different open-vocabulary queries. We use a diverse range of vocabulary categories to identify objects in scenes, such as visual attribute terms like ``green'', and subjective adjectives like ``cute''. Furthermore, the figure shows our method's effectiveness in identifying both large-scale objects like "lego" and specific components of an object, such as a "engine". We show more examples in the supplementary material.

\subsection{Ablation Study}
We demonstrate the results of ablation studies in Tab.~\ref{tab:ab} and Fig.~\ref{fig:abl}. 
The results show that embedding uncertainty without spatial smoothing of semantic features leads to suboptimal optimization. Conversely, using MLP solely for spatial smoothing, without accounting for semantic variance across multiple views, also impedes the optimization of precise language features. However, our full model, combining uncertainty with MLP smoothing in an adaptive manner effectively diminishes ambiguity and enhances the mean average precision (mAP) metric. 
Furthermore, integrating DINO features significantly improves the definition of object query boundaries. The load balancing loss, introduced during the quantization phase, results in a more utilized discrete feature space, facilitating the distinguish of objects with similar semantics, thereby boosting overall accuracy.

\section{Conclusion}
\label{sec:conclusion}

We present Language Embedded 3D Gaussians, a novel scene representation designed for open-vocabulary query tasks. Our method successfully embeds quantized compact semantics features onto massive 3D Gaussians, while only maintaining minimal memory and storage requirements. To address semantic inconsistencies across different viewpoints, we propose a feature smoothing procedure that adaptively lowers the spatial frequency of embedded semantic features, guided by the uncertainty values learned on the 3D Gaussians. The result is a highly effective scene representation that not only enables high-quality novel view synthesis but also provides high accuracy in open-vocabulary querying, all achieved with modest computational resources.

\paragraph{Limitations and Future works.}
Although DINO features improve object boundary detection, they fall short in pinpointing fine-grained object geometries at high resolutions when using CLIP-derived semantics. Additionally, detecting highly reflective or translucent objects, such as televisions and mirrors, remains challenging. These limitations might be overcome with more advanced visual-language models and native per-pixel semantic features.
Nevertheless, our approach can be further adapted for broader open-vocabulary tasks, such as editing and generation of semantic-level scene objects.
{
    \small
    \bibliographystyle{ieeenat_fullname}
    \bibliography{main}

\begin{thebibliography}{55}
\providecommand{\natexlab}[1]{#1}
\providecommand{\url}[1]{\texttt{#1}}
\expandafter\ifx\csname urlstyle\endcsname\relax
  \providecommand{\doi}[1]{doi: #1}\else
  \providecommand{\doi}{doi: \begingroup \urlstyle{rm}\Url}\fi

\bibitem[Amir et~al.(2021)Amir, Gandelsman, Bagon, and Dekel]{amir2021deep}
Shir Amir, Yossi Gandelsman, Shai Bagon, and Tali Dekel.
\newblock Deep vit features as dense visual descriptors.
\newblock \emph{arXiv preprint arXiv:2112.05814}, 2021.

\bibitem[Barron et~al.(2021)Barron, Mildenhall, Tancik, Hedman, Martin-Brualla, and Srinivasan]{barron2021mip}
Jonathan~T Barron, Ben Mildenhall, Matthew Tancik, Peter Hedman, Ricardo Martin-Brualla, and Pratul~P Srinivasan.
\newblock Mip-nerf: A multiscale representation for anti-aliasing neural radiance fields.
\newblock In \emph{Proceedings of the IEEE/CVF International Conference on Computer Vision}, pages 5855--5864, 2021.

\bibitem[Barron et~al.(2022)Barron, Mildenhall, Verbin, Srinivasan, and Hedman]{barron2022mip}
Jonathan~T Barron, Ben Mildenhall, Dor Verbin, Pratul~P Srinivasan, and Peter Hedman.
\newblock Mip-nerf 360: Unbounded anti-aliased neural radiance fields.
\newblock In \emph{Proceedings of the IEEE/CVF Conference on Computer Vision and Pattern Recognition}, pages 5470--5479, 2022.

\bibitem[Barron et~al.(2023)Barron, Mildenhall, Verbin, Srinivasan, and Hedman]{barron2023zip}
Jonathan~T Barron, Ben Mildenhall, Dor Verbin, Pratul~P Srinivasan, and Peter Hedman.
\newblock Zip-nerf: Anti-aliased grid-based neural radiance fields.
\newblock \emph{arXiv preprint arXiv:2304.06706}, 2023.

\bibitem[Caron et~al.(2021)Caron, Touvron, Misra, J\'egou, Mairal, Bojanowski, and Joulin]{caron2021emerging}
Mathilde Caron, Hugo Touvron, Ishan Misra, Herv\'e J\'egou, Julien Mairal, Piotr Bojanowski, and Armand Joulin.
\newblock Emerging properties in self-supervised vision transformers.
\newblock In \emph{Proceedings of the International Conference on Computer Vision (ICCV)}, 2021.

\bibitem[Chaurasia et~al.(2013)Chaurasia, Duchene, Sorkine-Hornung, and Drettakis]{chaurasia2013depth}
Gaurav Chaurasia, Sylvain Duchene, Olga Sorkine-Hornung, and George Drettakis.
\newblock Depth synthesis and local warps for plausible image-based navigation.
\newblock \emph{ACM Transactions on Graphics (TOG)}, 32\penalty0 (3), 2013.

\bibitem[Chen et~al.(2022)Chen, Xu, Geiger, Yu, and Su]{chen2022tensorf}
Anpei Chen, Zexiang Xu, Andreas Geiger, Jingyi Yu, and Hao Su.
\newblock Tensorf: Tensorial radiance fields.
\newblock In \emph{European Conference on Computer Vision}, pages 333--350. Springer, 2022.

\bibitem[Ding et~al.(2023)Ding, Yang, Xue, Zhang, Bai, and Qi]{ding2022language}
Runyu Ding, Jihan Yang, Chuhui Xue, Wenqing Zhang, Song Bai, and Xiaojuan Qi.
\newblock Pla: Language-driven open-vocabulary 3d scene understanding.
\newblock In \emph{Proceedings of the IEEE/CVF Conference on Computer Vision and Pattern Recognition}, 2023.

\bibitem[Eisemann et~al.(2008)Eisemann, De~Decker, Magnor, Bekaert, De~Aguiar, Ahmed, Theobalt, and Sellent]{eisemann2008floating}
Martin Eisemann, Bert De~Decker, Marcus Magnor, Philippe Bekaert, Edilson De~Aguiar, Naveed Ahmed, Christian Theobalt, and Anita Sellent.
\newblock Floating textures.
\newblock In \emph{Computer graphics forum}, pages 409--418. Wiley Online Library, 2008.

\bibitem[Engelmann et~al.(2023)Engelmann, Manhardt, Niemeyer, Tateno, Pollefeys, and Tombari]{engelmann2023open}
Francis Engelmann, Fabian Manhardt, Michael Niemeyer, Keisuke Tateno, Marc Pollefeys, and Federico Tombari.
\newblock Open-set 3d scene segmentation with rendered novel views.
\newblock 2023.

\bibitem[Fedus et~al.(2022)Fedus, Zoph, and Shazeer]{fedus2022switch}
William Fedus, Barret Zoph, and Noam Shazeer.
\newblock Switch transformers: Scaling to trillion parameter models with simple and efficient sparsity.
\newblock \emph{The Journal of Machine Learning Research}, 23\penalty0 (1):\penalty0 5232--5270, 2022.

\bibitem[Fridovich-Keil et~al.(2022)Fridovich-Keil, Yu, Tancik, Chen, Recht, and Kanazawa]{fridovich2022plenoxels}
Sara Fridovich-Keil, Alex Yu, Matthew Tancik, Qinhong Chen, Benjamin Recht, and Angjoo Kanazawa.
\newblock Plenoxels: Radiance fields without neural networks.
\newblock In \emph{Proceedings of the IEEE/CVF Conference on Computer Vision and Pattern Recognition}, pages 5501--5510, 2022.

\bibitem[Fu et~al.(2022)Fu, Zhang, Chen, Lu, Zhu, Zhou, Geiger, and Liao]{fu2022panoptic}
Xiao Fu, Shangzhan Zhang, Tianrun Chen, Yichong Lu, Lanyun Zhu, Xiaowei Zhou, Andreas Geiger, and Yiyi Liao.
\newblock Panoptic nerf: 3d-to-2d label transfer for panoptic urban scene segmentation.
\newblock In \emph{2022 International Conference on 3D Vision (3DV)}, pages 1--11. IEEE, 2022.

\bibitem[Goesele et~al.(2007)Goesele, Snavely, Curless, Hoppe, and Seitz]{goesele2007multi}
Michael Goesele, Noah Snavely, Brian Curless, Hugues Hoppe, and Steven~M Seitz.
\newblock Multi-view stereo for community photo collections.
\newblock In \emph{2007 IEEE 11th International Conference on Computer Vision}, pages 1--8. IEEE, 2007.

\bibitem[Hedman et~al.(2018)Hedman, Philip, Price, Frahm, Drettakis, and Brostow]{hedman2018deep}
Peter Hedman, Julien Philip, True Price, Jan-Michael Frahm, George Drettakis, and Gabriel Brostow.
\newblock Deep blending for free-viewpoint image-based rendering.
\newblock \emph{ACM Transactions on Graphics (ToG)}, 37\penalty0 (6):\penalty0 1--15, 2018.

\bibitem[Huang et~al.(2022)Huang, Dong, Yang, Huang, Lau, Ouyang, and Zuo]{huang2022clip2point}
Tianyu Huang, Bowen Dong, Yunhan Yang, Xiaoshui Huang, Rynson~WH Lau, Wanli Ouyang, and Wangmeng Zuo.
\newblock Clip2point: Transfer clip to point cloud classification with image-depth pre-training.
\newblock \emph{arXiv preprint arXiv:2210.01055}, 2022.

\bibitem[Kerbl et~al.(2023)Kerbl, Kopanas, Leimk{\"u}hler, and Drettakis]{kerbl20233d}
Bernhard Kerbl, Georgios Kopanas, Thomas Leimk{\"u}hler, and George Drettakis.
\newblock 3d gaussian splatting for real-time radiance field rendering.
\newblock \emph{ACM Transactions on Graphics (ToG)}, 42\penalty0 (4):\penalty0 1--14, 2023.

\bibitem[Kerr et~al.(2023)Kerr, Kim, Goldberg, Kanazawa, and Tancik]{kerr2023lerf}
Justin Kerr, Chung~Min Kim, Ken Goldberg, Angjoo Kanazawa, and Matthew Tancik.
\newblock Lerf: Language embedded radiance fields.
\newblock In \emph{Proceedings of the IEEE/CVF International Conference on Computer Vision}, pages 19729--19739, 2023.

\bibitem[Kobayashi et~al.(2022)Kobayashi, Matsumoto, and Sitzmann]{kobayashi2022decomposing}
Sosuke Kobayashi, Eiichi Matsumoto, and Vincent Sitzmann.
\newblock Decomposing nerf for editing via feature field distillation.
\newblock \emph{Advances in Neural Information Processing Systems}, 35:\penalty0 23311--23330, 2022.

\bibitem[Kundu et~al.(2022)Kundu, Genova, Yin, Fathi, Pantofaru, Guibas, Tagliasacchi, Dellaert, and Funkhouser]{kundu2022panoptic}
Abhijit Kundu, Kyle Genova, Xiaoqi Yin, Alireza Fathi, Caroline Pantofaru, Leonidas~J Guibas, Andrea Tagliasacchi, Frank Dellaert, and Thomas Funkhouser.
\newblock Panoptic neural fields: A semantic object-aware neural scene representation.
\newblock In \emph{Proceedings of the IEEE/CVF Conference on Computer Vision and Pattern Recognition}, pages 12871--12881, 2022.

\bibitem[Li et~al.(2022)Li, Weinberger, Belongie, Koltun, and Ranftl]{li2022languagedriven}
Boyi Li, Kilian~Q Weinberger, Serge Belongie, Vladlen Koltun, and Rene Ranftl.
\newblock Language-driven semantic segmentation.
\newblock In \emph{International Conference on Learning Representations}, 2022.

\bibitem[Liu et~al.(2023{\natexlab{a}})Liu, Zhan, Zhang, Xu, Yu, Saddik, Theobalt, Xing, and Lu]{liu20233d}
Kunhao Liu, Fangneng Zhan, Jiahui Zhang, Muyu Xu, Yingchen Yu, Abdulmotaleb~El Saddik, Christian Theobalt, Eric Xing, and Shijian Lu.
\newblock 3d open-vocabulary segmentation with foundation models.
\newblock \emph{arXiv preprint arXiv:2305.14093}, 2023{\natexlab{a}}.

\bibitem[Liu et~al.(2023{\natexlab{b}})Liu, Zhan, Zhang, Xu, Yu, Saddik, Theobalt, Xing, and Lu]{liu2023weakly}
Kunhao Liu, Fangneng Zhan, Jiahui Zhang, Muyu Xu, Yingchen Yu, Abdulmotaleb~El Saddik, Christian Theobalt, Eric Xing, and Shijian Lu.
\newblock Weakly supervised 3d open-vocabulary segmentation.
\newblock \emph{arXiv preprint arXiv:2305.14093}, 2023{\natexlab{b}}.

\bibitem[Liu et~al.(2020)Liu, Gu, Zaw~Lin, Chua, and Theobalt]{liu2020neural}
Lingjie Liu, Jiatao Gu, Kyaw Zaw~Lin, Tat-Seng Chua, and Christian Theobalt.
\newblock Neural sparse voxel fields.
\newblock \emph{Advances in Neural Information Processing Systems}, 33:\penalty0 15651--15663, 2020.

\bibitem[Loshchilov and Hutter(2019)]{loshchilov2019decoupled}
Ilya Loshchilov and Frank Hutter.
\newblock Decoupled weight decay regularization, 2019.

\bibitem[Lu et~al.(2023)Lu, Xu, Wei, Xie, Tomizuka, Keutzer, and Zhang]{lu2023open}
Yuheng Lu, Chenfeng Xu, Xiaobao Wei, Xiaodong Xie, Masayoshi Tomizuka, Kurt Keutzer, and Shanghang Zhang.
\newblock Open-vocabulary point-cloud object detection without 3d annotation.
\newblock 2023.

\bibitem[Martin-Brualla et~al.(2021)Martin-Brualla, Radwan, Sajjadi, Barron, Dosovitskiy, and Duckworth]{martin2021nerf}
Ricardo Martin-Brualla, Noha Radwan, Mehdi~SM Sajjadi, Jonathan~T Barron, Alexey Dosovitskiy, and Daniel Duckworth.
\newblock Nerf in the wild: Neural radiance fields for unconstrained photo collections.
\newblock In \emph{Proceedings of the IEEE/CVF Conference on Computer Vision and Pattern Recognition}, pages 7210--7219, 2021.

\bibitem[Mildenhall et~al.(2021)Mildenhall, Srinivasan, Tancik, Barron, Ramamoorthi, and Ng]{mildenhall2021nerf}
Ben Mildenhall, Pratul~P Srinivasan, Matthew Tancik, Jonathan~T Barron, Ravi Ramamoorthi, and Ren Ng.
\newblock Nerf: Representing scenes as neural radiance fields for view synthesis.
\newblock \emph{Communications of the ACM}, 65\penalty0 (1):\penalty0 99--106, 2021.

\bibitem[M{\"u}ller et~al.(2022)M{\"u}ller, Evans, Schied, and Keller]{muller2022instant}
Thomas M{\"u}ller, Alex Evans, Christoph Schied, and Alexander Keller.
\newblock Instant neural graphics primitives with a multiresolution hash encoding.
\newblock \emph{ACM Transactions on Graphics (ToG)}, 41\penalty0 (4):\penalty0 1--15, 2022.

\bibitem[Paszke et~al.(2017)Paszke, Gross, Chintala, Chanan, Yang, DeVito, Lin, Desmaison, Antiga, and Lerer]{paszke2017automatic}
Adam Paszke, Sam Gross, Soumith Chintala, Gregory Chanan, Edward Yang, Zachary DeVito, Zeming Lin, Alban Desmaison, Luca Antiga, and Adam Lerer.
\newblock Automatic differentiation in pytorch.
\newblock In \emph{NIPS-W}, 2017.

\bibitem[Peng et~al.(2023)Peng, Genova, Jiang, Tagliasacchi, Pollefeys, and Funkhouser]{Peng2023OpenScene}
Songyou Peng, Kyle Genova, Chiyu~"Max" Jiang, Andrea Tagliasacchi, Marc Pollefeys, and Thomas Funkhouser.
\newblock Openscene: 3d scene understanding with open vocabularies.
\newblock In \emph{Proceedings of the IEEE/CVF Conference on Computer Vision and Pattern Recognition (CVPR)}, 2023.

\bibitem[Radford et~al.(2021)Radford, Kim, Hallacy, Ramesh, Goh, Agarwal, Sastry, Askell, Mishkin, Clark, et~al.]{radford2021learning}
Alec Radford, Jong~Wook Kim, Chris Hallacy, Aditya Ramesh, Gabriel Goh, Sandhini Agarwal, Girish Sastry, Amanda Askell, Pamela Mishkin, Jack Clark, et~al.
\newblock Learning transferable visual models from natural language supervision.
\newblock In \emph{International conference on machine learning}, pages 8748--8763. PMLR, 2021.

\bibitem[Rematas et~al.(2022)Rematas, Liu, Srinivasan, Barron, Tagliasacchi, Funkhouser, and Ferrari]{rematas2022urf}
Konstantinos Rematas, Andrew Liu, Pratul~P. Srinivasan, Jonathan~T. Barron, Andrea Tagliasacchi, Tom Funkhouser, and Vittorio Ferrari.
\newblock Urban radiance fields.
\newblock \emph{CVPR}, 2022.

\bibitem[Siddiqui et~al.(2023)Siddiqui, Porzi, Bul{\`o}, M{\"u}ller, Nie{\ss}ner, Dai, and Kontschieder]{siddiqui2023panoptic}
Yawar Siddiqui, Lorenzo Porzi, Samuel~Rota Bul{\`o}, Norman M{\"u}ller, Matthias Nie{\ss}ner, Angela Dai, and Peter Kontschieder.
\newblock Panoptic lifting for 3d scene understanding with neural fields.
\newblock In \emph{Proceedings of the IEEE/CVF Conference on Computer Vision and Pattern Recognition}, pages 9043--9052, 2023.

\bibitem[Snavely et~al.(2006)Snavely, Seitz, and Szeliski]{snavely2006photo}
Noah Snavely, Steven~M Seitz, and Richard Szeliski.
\newblock Photo tourism: exploring photo collections in 3d.
\newblock In \emph{ACM siggraph 2006 papers}, pages 835--846. 2006.

\bibitem[Sun et~al.(2022)Sun, Sun, and Chen]{sun2022direct}
Cheng Sun, Min Sun, and Hwann-Tzong Chen.
\newblock Direct voxel grid optimization: Super-fast convergence for radiance fields reconstruction.
\newblock In \emph{Proceedings of the IEEE/CVF Conference on Computer Vision and Pattern Recognition}, pages 5459--5469, 2022.

\bibitem[Takmaz et~al.(2023)Takmaz, Fedele, Sumner, Pollefeys, Tombari, and Engelmann]{takmaz2023openmask3d}
Ay{\c{c}}a Takmaz, Elisabetta Fedele, Robert~W. Sumner, Marc Pollefeys, Federico Tombari, and Francis Engelmann.
\newblock {OpenMask3D: Open-Vocabulary 3D Instance Segmentation}.
\newblock In \emph{Advances in Neural Information Processing Systems (NeurIPS)}, 2023.

\bibitem[Tancik et~al.(2022)Tancik, Casser, Yan, Pradhan, Mildenhall, Srinivasan, Barron, and Kretzschmar]{Tancik_2022_CVPR}
Matthew Tancik, Vincent Casser, Xinchen Yan, Sabeek Pradhan, Ben Mildenhall, Pratul~P. Srinivasan, Jonathan~T. Barron, and Henrik Kretzschmar.
\newblock Block-nerf: Scalable large scene neural view synthesis.
\newblock In \emph{Proceedings of the IEEE/CVF Conference on Computer Vision and Pattern Recognition (CVPR)}, pages 8248--8258, 2022.

\bibitem[Tschernezki et~al.(2022)Tschernezki, Laina, Larlus, and Vedaldi]{tschernezki2022neural}
Vadim Tschernezki, Iro Laina, Diane Larlus, and Andrea Vedaldi.
\newblock Neural feature fusion fields: 3d distillation of self-supervised 2d image representations.
\newblock In \emph{2022 International Conference on 3D Vision (3DV)}, pages 443--453. IEEE, 2022.

\bibitem[Van Den~Oord et~al.(2017)Van Den~Oord, Vinyals, et~al.]{van2017neural}
Aaron Van Den~Oord, Oriol Vinyals, et~al.
\newblock Neural discrete representation learning.
\newblock \emph{Advances in neural information processing systems}, 30, 2017.

\bibitem[Wang et~al.(2022)Wang, Chen, and Yang]{wang2022dm}
Bing Wang, Lu Chen, and Bo Yang.
\newblock Dm-nerf: 3d scene geometry decomposition and manipulation from 2d images.
\newblock \emph{arXiv preprint arXiv:2208.07227}, 2022.

\bibitem[Wang et~al.(2023)Wang, He, Peng, Lin, Bao, and Zhou]{wang2023autorecon}
Yuang Wang, Xingyi He, Sida Peng, Haotong Lin, Hujun Bao, and Xiaowei Zhou.
\newblock Autorecon: Automated 3d object discovery and reconstruction.
\newblock In \emph{CVPR}, 2023.

\bibitem[Wu et~al.(2023)Wu, Li, Xu, Yuan, Ding, Yang, Li, Zhang, Tong, Jiang, Ghanem, and Tao]{wu2023open}
Jianzong Wu, Xiangtai Li, Shilin Xu, Haobo Yuan, Henghui Ding, Yibo Yang, Xia Li, Jiangning Zhang, Yunhai Tong, Xudong Jiang, Bernard Ghanem, and Dacheng Tao.
\newblock Towards open vocabulary learning: A survey.
\newblock \emph{arXiv pre-print}, 2023.

\bibitem[Xu et~al.(2022)Xu, Xu, Philip, Bi, Shu, Sunkavalli, and Neumann]{xu2022point}
Qiangeng Xu, Zexiang Xu, Julien Philip, Sai Bi, Zhixin Shu, Kalyan Sunkavalli, and Ulrich Neumann.
\newblock Point-nerf: Point-based neural radiance fields.
\newblock In \emph{Proceedings of the IEEE/CVF Conference on Computer Vision and Pattern Recognition}, pages 5438--5448, 2022.

\bibitem[Xue et~al.(2022)Xue, Gao, Xing, Mart{\'\i}n-Mart{\'\i}n, Wu, Xiong, Xu, Niebles, and Savarese]{xue2022ulip}
Le Xue, Mingfei Gao, Chen Xing, Roberto Mart{\'\i}n-Mart{\'\i}n, Jiajun Wu, Caiming Xiong, Ran Xu, Juan~Carlos Niebles, and Silvio Savarese.
\newblock Ulip: Learning unified representation of language, image and point cloud for 3d understanding.
\newblock \emph{arXiv preprint arXiv:2212.05171}, 2022.

\bibitem[Xue et~al.(2023)Xue, Yu, Zhang, Li, Martín-Martín, Wu, Xiong, Xu, Niebles, and Savarese]{xue2023ulip2}
Le Xue, Ning Yu, Shu Zhang, Junnan Li, Roberto Martín-Martín, Jiajun Wu, Caiming Xiong, Ran Xu, Juan~Carlos Niebles, and Silvio Savarese.
\newblock Ulip-2: Towards scalable multimodal pre-training for 3d understanding, 2023.

\bibitem[Yang et~al.(2023{\natexlab{a}})Yang, Chen, Qian, Fouhey, and Chai]{chat-with-nerf-2023}
Jianing Yang, Xuweiyi Chen, Shengyi Qian, David Fouhey, and Joyce Chai.
\newblock Chat with nerf: Grounding 3d objects in neural radiance field through dialog, 2023{\natexlab{a}}.

\bibitem[Yang et~al.(2023{\natexlab{b}})Yang, Chen, Qian, Madaan, Iyengar, Fouhey, and Chai]{yang2023llmgrounder}
Jianing Yang, Xuweiyi Chen, Shengyi Qian, Nikhil Madaan, Madhavan Iyengar, David~F. Fouhey, and Joyce Chai.
\newblock Llm-grounder: Open-vocabulary 3d visual grounding with large language model as an agent, 2023{\natexlab{b}}.

\bibitem[Zhang et~al.(2023{\natexlab{a}})Zhang, Dong, and Ma]{zhang2023clipfo3d}
Junbo Zhang, Runpei Dong, and Kaisheng Ma.
\newblock Clip-fo3d: Learning free open-world 3d scene representations from 2d dense clip, 2023{\natexlab{a}}.

\bibitem[Zhang et~al.(2020)Zhang, Riegler, Snavely, and Koltun]{zhang2020nerf++}
Kai Zhang, Gernot Riegler, Noah Snavely, and Vladlen Koltun.
\newblock Nerf++: Analyzing and improving neural radiance fields.
\newblock \emph{arXiv preprint arXiv:2010.07492}, 2020.

\bibitem[Zhang et~al.(2018)Zhang, Isola, Efros, Shechtman, and Wang]{zhang2018unreasonable}
Richard Zhang, Phillip Isola, Alexei~A. Efros, Eli Shechtman, and Oliver Wang.
\newblock The unreasonable effectiveness of deep features as a perceptual metric, 2018.

\bibitem[Zhang et~al.(2021)Zhang, Guo, Zhang, Li, Miao, Cui, Qiao, Gao, and Li]{zhang2021pointclip}
Renrui Zhang, Ziyu Guo, Wei Zhang, Kunchang Li, Xupeng Miao, Bin Cui, Yu Qiao, Peng Gao, and Hongsheng Li.
\newblock Pointclip: Point cloud understanding by clip.
\newblock \emph{arXiv preprint arXiv:2112.02413}, 2021.

\bibitem[Zhang et~al.(2023{\natexlab{b}})Zhang, Kundu, Funkhouser, Guibas, Su, and Genova]{zhang2023nerflets}
Xiaoshuai Zhang, Abhijit Kundu, Thomas Funkhouser, Leonidas Guibas, Hao Su, and Kyle Genova.
\newblock Nerflets: Local radiance fields for efficient structure-aware 3d scene representation from 2d supervision.
\newblock In \emph{Proceedings of the IEEE/CVF Conference on Computer Vision and Pattern Recognition}, pages 8274--8284, 2023{\natexlab{b}}.

\bibitem[Zhi et~al.(2021)Zhi, Laidlow, Leutenegger, and Davison]{zhi2021place}
Shuaifeng Zhi, Tristan Laidlow, Stefan Leutenegger, and Andrew~J Davison.
\newblock In-place scene labelling and understanding with implicit scene representation.
\newblock In \emph{Proceedings of the IEEE/CVF International Conference on Computer Vision}, pages 15838--15847, 2021.

\bibitem[Zhu et~al.(2022)Zhu, Zhang, He, Guo, Zeng, Qin, Zhang, and Gao]{Zhu2022PointCLIPV2}
Xiangyang Zhu, Renrui Zhang, Bowei He, Ziyu Guo, Ziyao Zeng, Zipeng Qin, Shanghang Zhang, and Peng Gao.
\newblock Pointclip v2: Prompting clip and gpt for powerful 3d open-world learning.
\newblock \emph{arXiv preprint arXiv:2211.11682}, 2022.

\end{thebibliography}
}
\clearpage
\setcounter{page}{1}
\maketitlesupplementary

We provide more details in this supplementary document, including load balancing loss (Sec.~\ref{sec:loss}), inference strategy (Sec.~\ref{sec:infer}), implementation details (Sec.~\ref{sec:impl}), 
dataset details (Sec.~\ref{sec:data}) and more Results (Sec.~\ref{sec:res}).

\section{Load Balancing Loss}
\label{sec:loss}
To maximize utilization of the optimized feature space and avert quantization collapse, we introduce a load balancing loss, inspired by Switch Transformer~\cite{fedus2022switch}.
When quantizing $K$ features, the utilization ratio of each feature in $\mathcal{S}$ is calculated as:
\begin{equation}
    m_i = {\text{argmax}}_{j}(\mathcal{D}(\mathbf{F}_i, \mathbf{f}_j)), \, \text{where} \, \mathbf{f}_j \in \mathcal{S}, 
\end{equation}
\begin{equation}
    \mathbf{r} = \frac {\sum^K_i \text{onehot}(m_i)} {K},
\end{equation}
where $\mathbf{r} \in \mathbb{R}^N$.
We compute the mean selection probability for each feature over K quantizations:
\begin{equation}
    \mathcal{D}(\mathbf{F}_i, \mathcal{S}) = [\mathcal{D}(\mathbf{F}_i, \mathbf{f}_1), \mathcal{D}(\mathbf{F}_i, \mathbf{f}_2), \ldots, \mathcal{D}(\mathbf{F}_i, \mathbf{f}_N)],
\end{equation}
\begin{equation}
    \mathbf{p} = 
    \frac{\sum^K_i\text{Softmax}(\mathcal{D}(\mathbf{F}_i, \mathcal{S}))}{K},
\end{equation}
where $\mathcal{D}(\mathbf{F}_i, \mathcal{S}) \in \mathbb{R}^N$ and $\mathbf{p} \in \mathbb{R}^N$.
The load balancing loss is then computed by the element-wise multiplication of $\mathbf{r}$ and $\mathbf{p}$, followed by their aggregation:
\begin{equation}
    \mathcal{L}_{lb} = \sum^N(\mathbf{r} \circ \mathbf{p}),
\end{equation}
where $\circ$ denotes the element-wise product.

\section{Inference Strategy}
\label{sec:infer}
\begin{figure}[t] \centering
\includegraphics[width=0.5\textwidth]{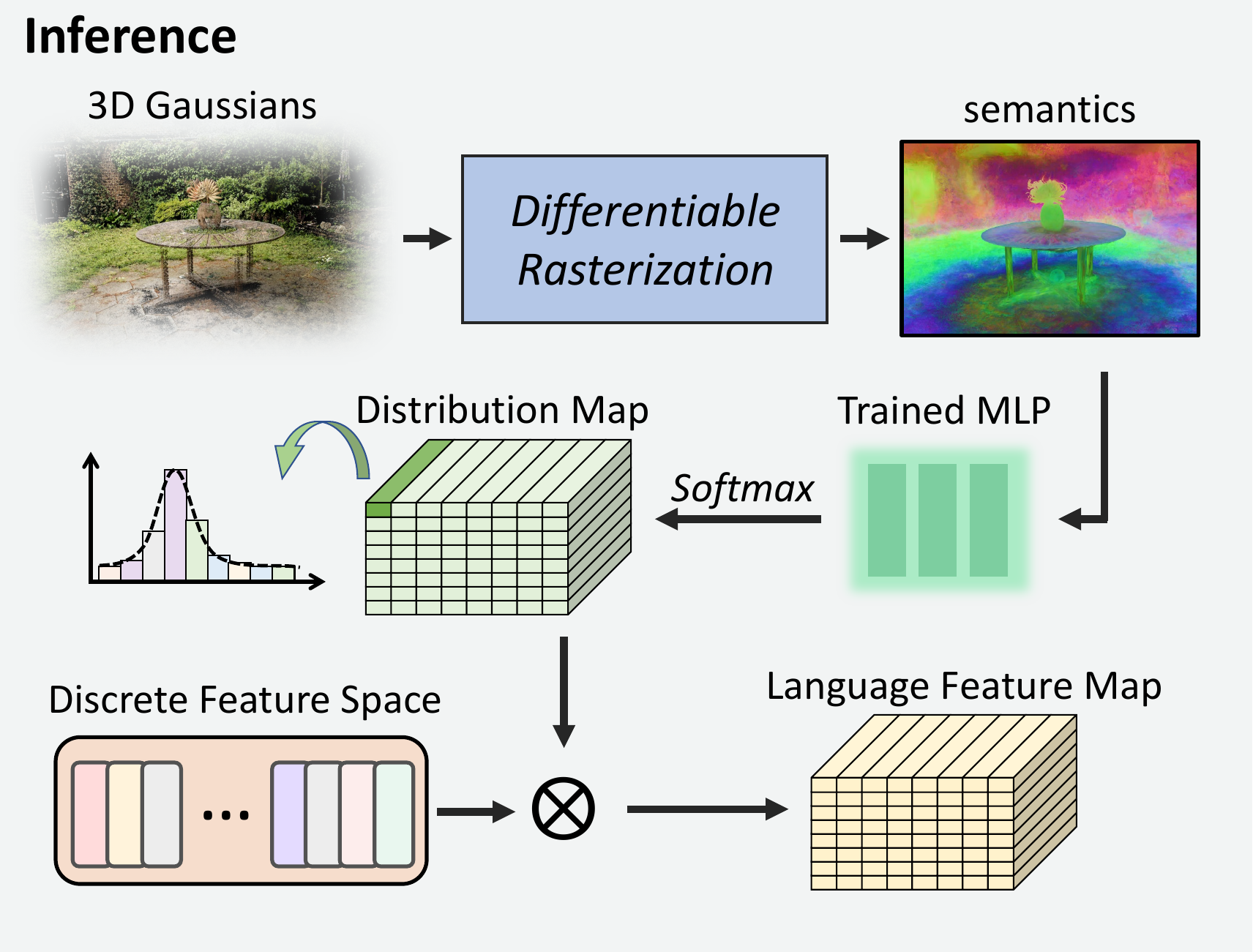}
    \caption{
    The inference pipeline for language feature maps from optimized 3D Gaussians.
    Each element in the distribution map $\mathcal{M}_{\text{infer}}$ represents a probability distribution of features in the Discrete Feature Space $\mathcal{S}$. 
    We compute the corresponding language feature map $\mathcal{F}$ based on $\mathcal{S}$ utilizing these distributions.
    }
    \label{fig:pipe}
\end{figure}

In the inference stage, rasterization and alpha blending are employed to project the compact semantic features of 3D Gaussians into a 2D feature map. 
This feature map is then converted into the distribution of semantic indices using a trained MLP decoder and softmax activation, expressed as:
\begin{equation}
    \mathcal{M}_{\text{infer}} = \text{Softmax}(D_{\text{MLP}}(R_s(\mathcal{G};p_{\text{cam}}))) ,
\end{equation}
where $ R_s(\mathcal{G};p_{\text{cam}}) \in \mathbb{R}^{H \times W \times d_{s}} $ denotes the rendered semantic features from a set of 3D Gaussians $ \mathcal{G} $, as observed from the camera pose $ p_{\text{cam}} $. Here, $ D_{\text{MLP}} $ symbolizes the trained MLP decoder of semantic features on 3D Gaussians.
The result is a language feature index distribution $ \mathcal{M}_{\text{infer}} \in \mathbb{R}^{H \times W \times N} $, where $ H $ and $ W $ represent the image's height and width, respectively. 
We finally acquire the language feature map, by multiplying $\mathcal{M}_{\text{infer}}$ with the quantized language features matrix $\mathbf{S} \in \mathbb{R}^{N \times d}$:
\begin{equation}
    \mathcal{F} = \mathcal{M}_{\text{infer}}\mathbf{S} ,
\end{equation}
where $\mathcal{F} \in \mathbb{R}^{H \times W \times d}$ denotes the language feature map derived from the 3D Gaussians $ \mathcal{G} $ observed from the camera pose $p_{\text{cam}}$. 
Figure~\ref{fig:pipe} depicts the inference pipeline for generating language feature maps from language-embedded 3D Gaussians.

Utilizing the provided text prompt, we identify objects within the 3D scene by computing the relevance map of $\mathcal{F}$ in accordance with LERF~\cite{kerr2023lerf}.

\begin{table}[t]\centering
    \resizebox{1.0\columnwidth}{!}{
    \footnotesize
    \begin{tabular}{*{4}{c}}
        \toprule
        & Layer & Config & Out Size \\
        \midrule
        Input & - & - & $8 \times H \times W$ \\
        C1 & Conv+ReLU & $128 \times 1 \times 1 \, / \, 1$ & $128 \times H \times W$ \\
        C2 & Conv+ReLU & $256 \times 1 \times 1 \, / \, 1$ & $256 \times H \times W$ \\
        C3 & Conv & $N \times 1 \times 1 \, / \, 1$ & $N \times H \times W$  \\
        \bottomrule
    \end{tabular}
    }
    \caption{Details of our semantic feature decoder $D_{\text{MLP}}$. In a layer characterized by $ c \times w \times w / s $, $ c $ represents the number of filters, $ w \times w $ indicates the filter size, and $ s $ denotes the stride size. The output dimensionality is expressed in terms of channel $\times$ height $\times$ width. The dimension of semantic feature on 3D Gaussians is $8$ and $N$ represents the size of discrete language feature space $\mathcal{S}$.}
    \label{tab:dmlp}
\end{table}

\begin{table}[t]\centering
    \resizebox{1.0\columnwidth}{!}{
    \footnotesize
    \begin{tabular}{*{4}{c}}
        \toprule
        & Layer & Config & Out Size \\
        \midrule
        PE & Positional Encoding & $0$ & $3$ \\
        F1 & Full-connected+ReLU & $128$ & $128$ \\
        F2 & Full-connected+ReLU & $128$ & $128$ \\
        F3 & Full-connected+ReLU & $128$ & $128$ \\
        F4 & Full-connected & $8$ & $8$ \\
        \bottomrule
    \end{tabular}
    }
    \caption{Details of the PE and MLP in the adaptive spatial smoothing. The input is the position of 3D Gaussian and the output is smoothed semantic feature $s_{\text{MLP}}$. The configure of PE layer represents the frequency of positional encoding.}
    \label{tab:pemlp}
\end{table}

\section{Implementation Details}
\label{sec:impl}
In Tab.~\ref{tab:dmlp} and Tab.~\ref{tab:pemlp}, we present the implementation details of the semantic feature decoder $ D_{\text{MLP}} $ and the PE and MLP in the adaptive spatial smoothing, respectively. 
Furthermore, for the size $ N $ of the discrete language feature space $ \mathcal{S} $, we set $ N  = 32 $ for the "kitchen" scene, $ N  = 64 $ for the "bonsai" scene, and $ N  = 128 $ for other scenes. 
$ N $, as a hyperparameter, controls the capacity of semantic information in the discrete language feature space $ \mathcal{S} $ and can be adjusted according to the richness of semantic information in the scene.

\section{Datasets}
\label{sec:data}
\begin{figure*}[t] \centering
    \includegraphics[width=0.9\textwidth]{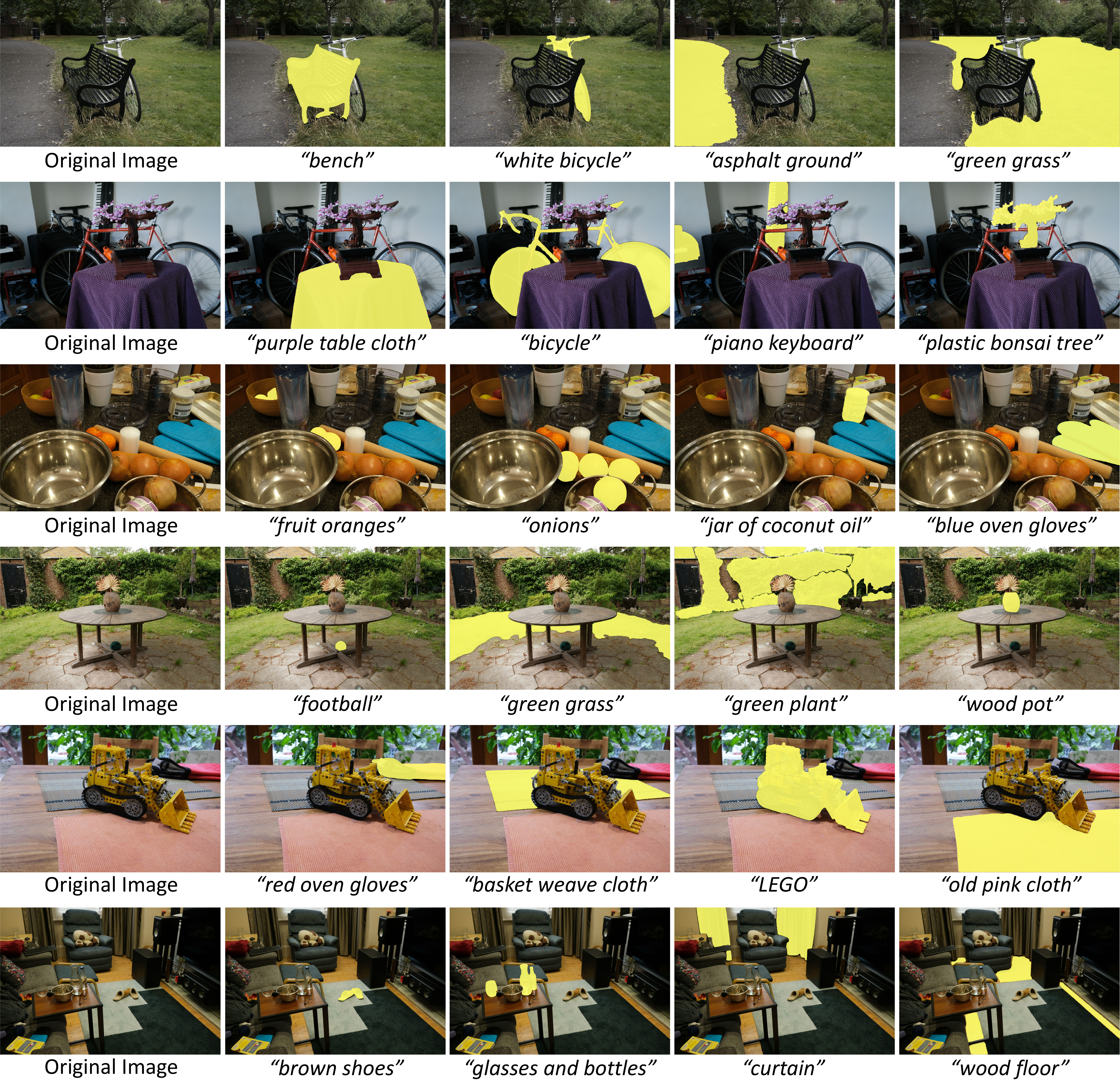}
    \caption{
    Ground truth segmentation masks for some objects in our dataset.
    In each scene, we select primary, unambiguous objects for semantic annotation. This includes both large and small objects, as well as challenging entities with complex geometric structures or transparent and translucent properties, such as bicycles, windows, and water glasses.
    }
    \label{fig:data}
\end{figure*}

To concurrently assess the quality of visual and semantic embeddings, six scenes from the Mip-NeRF360 dataset~\cite{barron2022mip} are chosen for quantitative and qualitative evaluation. 
The `Stump' scene is excluded due to its insufficient semantic content. 
The evaluation set of each scene is manually annotated with segmentation maps, which are created for the primary objects in each scene.
The text prompts corresponding to these annotated objects are listed in Tab.~\ref{tab:prompt}. 
Additionally, segmentation masks for some objects in our dataset are illustrated in Fig.~\ref{fig:data}.

\section{More Results}
\label{sec:res}
\begin{figure*}[t] \centering
    \includegraphics[width=0.92\textwidth]{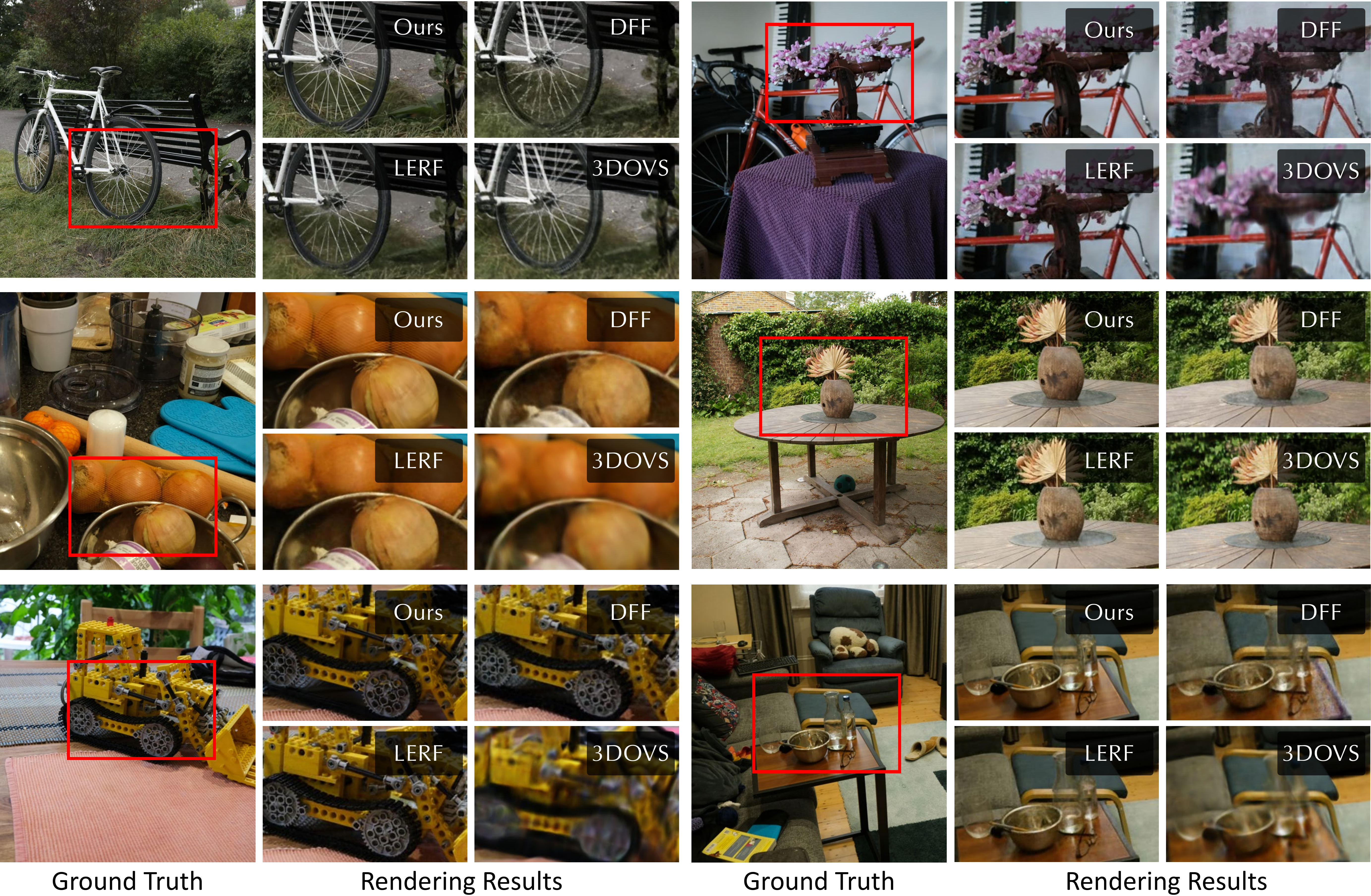}
    \caption{
    Comparison of the quality of novel view synthesis. Even with dense language features embedded into the 3D Gaussians, our method still only requires a reasonable amount of memory, thus allowing a massive amount of points to be rendered and optimized at the same time, achieving the best visual quality with more details compared to other methods.
    }
    \label{fig:visual}
    \vspace{-2.0mm}
\end{figure*}

\begin{figure*}[t] \centering
    \includegraphics[width=0.92\textwidth]{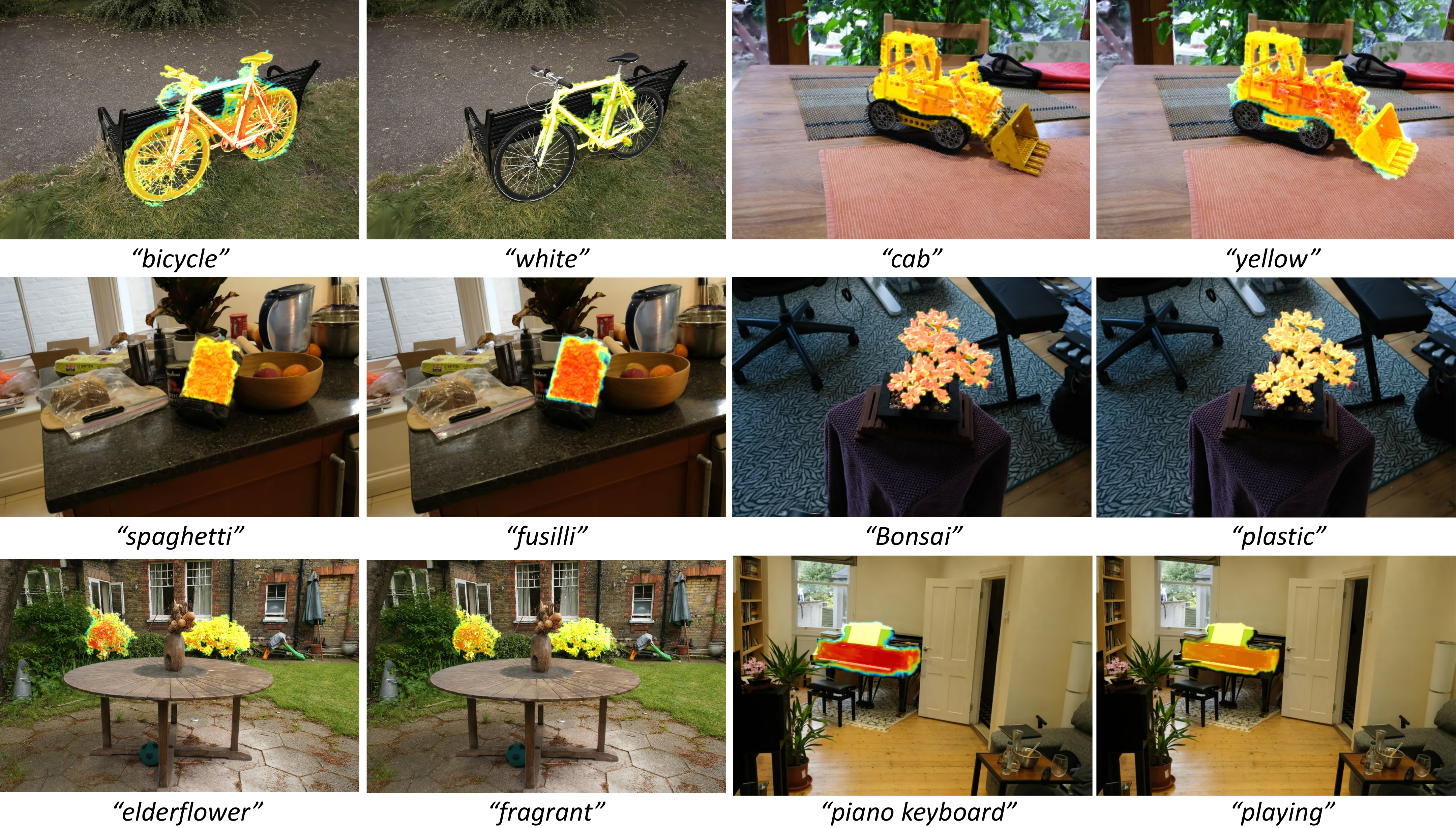}
    \caption{
    Examples of various open-vocabulary queries. 
    Our approach enables accurate open-vocabulary queries using a diverse class of word types, including but not limited to, visual attributes, general terms, materials, olfactory properties, and related actions.
    }
    \label{fig:query}
    \vspace{-2.0mm}
\end{figure*}

Further qualitative results are presented to illustrate the comparison of visual quality (Fig.~\ref{fig:visual}), the evaluation of novel view synthesis and query accuracy (Fig.~\ref{fig:comp}), and the exploration of open-vocabulary queries (Fig.~\ref{fig:query}).

\begin{table*}[t]\centering
\resizebox{0.9\textwidth}{!}{
\begin{tabular}{cc}
\toprule
Scene   & Positive Words \\ 
\midrule
bicycle & green grass, white bicycle, tire, bench, asphalt ground, silver oak tree \\
\midrule
bonsai  & piano keyboard, bicycle, purple table cloth, black stool, plastic bonsai tree, dark grey patterned carpet \\
\midrule
counter &
  \begin{tabular}[c]{@{}c@{}}jar of coconut oil, fruit oranges, onions, plants, blue oven gloves, wood rolling pin, \\ free range eggs box, stable bread, garofalo pasta, napolina tomatoes, gold ripple baking pan\end{tabular} \\
\midrule
garden  & football, wood round table, green grass, wood pot, elderflower, green plant, bricks wall, windows, stone ground \\
\midrule
kitchen & LEGO Technic 856 Bulldozer, basket weave cloth, wood plat, old pink striped cloth, red oven gloves \\
\midrule
room &
  \begin{tabular}[c]{@{}c@{}}blue grey chair, curtain, brown shoes, books, windows, door, \\ piano keyboard, wood floor, wine glasses and bottles, yucca plant, deep green carpets\end{tabular} \\
\bottomrule
\end{tabular}
}
\caption{Text prompts used for evaluating quality and accuracy of open-vocabulary query.}
\label{tab:prompt}
\end{table*}

\begin{figure*}[t] \centering
    \includegraphics[width=\textwidth]{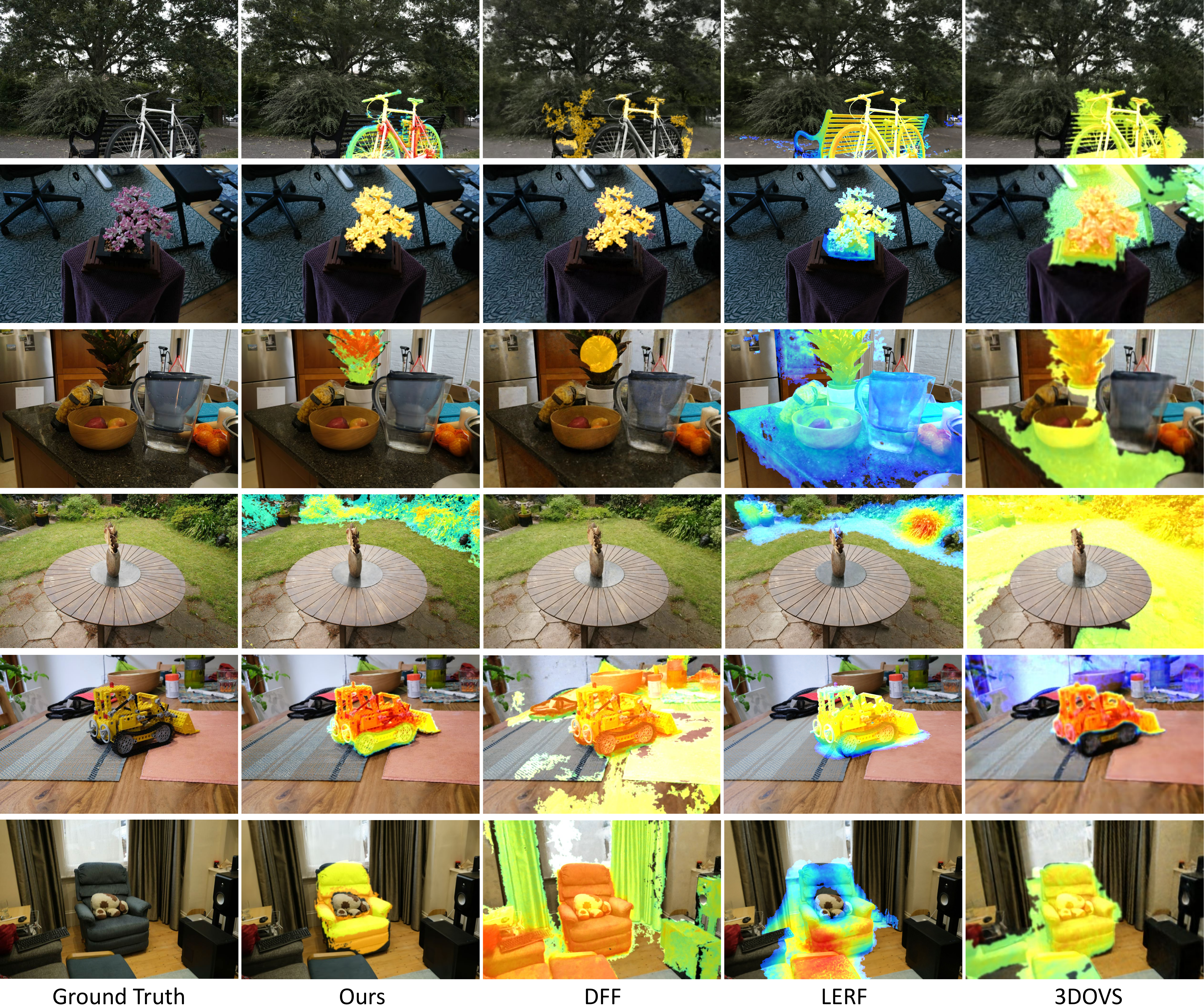}
    \caption{
    Comparison of novel view synthesis quality and open-vocabulary query accuracy. Left to right: Ground truth novel view synthesis, novel view images with relevance visualization from our method, DFF~\cite{kobayashi2022decomposing}, LeRF~\cite{kerr2023lerf}, and 3DOVS~\cite{liu20233d}. Top to bottom: Query words ``white bicycle'', ``bonsai'', ``plants'', ``green plant'', ``LEGO Technic 856 Bulldozer'', and ``blue grey chair''.
    }
    \label{fig:comp}
\end{figure*}

\end{document}